\documentclass[10pt,twocolumn,letterpaper]{article}

\usepackage{iccv}              
\usepackage{algorithm}
\usepackage{algpseudocode}
 \usepackage{multirow}
 \usepackage[accsupp]{axessibility}
\DeclareUnicodeCharacter{FF0C}{,}

\definecolor{iccvblue}{rgb}{0.21,0.49,0.74}
\usepackage[pagebackref,breaklinks,colorlinks,allcolors=iccvblue]{hyperref}
\usepackage[utf8]{inputenc}  

\usepackage{multicol}
\usepackage{pifont}
\usepackage{marvosym}
\def\paperID{4599} 
\def\confName{ICCV}
\def\confYear{2025}

\begin{document}

\title{X2I: Seamless Integration of Multimodal Understanding into Diffusion Transformer via Attention Distillation}

\author{Jian Ma \Letter\\
OPPO AI Center\\
{\tt\small majian2@oppo.com}
\and
Qirong Peng\thanks{Co-first authors}\\
OPPO AI Center\\
{\tt\small pengjirong@oppo.com}
\and
Xu Guo\thanks{The author did his work during internship at OPPO AI Center.}\\
Tsinghua University\\
{\tt\small guo-x24@mails.tsinghua.edu.cn}
\and
Chen Chen \Letter\\
OPPO AI Center\\
{\tt\small chenchen4@oppo.com}
\and
Haonan Lu\\
OPPO AI Center\\
{\tt\small luhaonan@oppo.com}
\and
Zhenyu Yang\\
OPPO AI Center\\
{\tt\small yangzhenyu@oppo.com}
}

\maketitle

\begin{abstract}
Text-to-image (T2I) models are well known for their ability to produce highly realistic images, while multimodal large language models (MLLMs) are renowned for their proficiency in understanding and integrating multiple modalities. However, currently there is no straightforward and efficient framework to transfer the multimodal comprehension abilities of MLLMs to T2I models to enable them to understand multimodal inputs. In this paper, we propose the X2I framework, which endows Diffusion Transformer (DiT) models with the capability to comprehend various modalities, including multilingual text, screenshot documents, images, videos, and audio. X2I is trained on a 100K English corpus in 160 GPU hours. Building on the DiT teacher model, we adopt an innovative distillation method to extract the inference capabilities of the teacher model and design a lightweight AlignNet structure to serve as an intermediate bridge. Compared to the teacher model, X2I shows a decrease in performance degradation of less than 1\% while gaining various multimodal understanding abilities, including multilingual to image, image to image, image-text to image, video to image, audio to image, and utilizing creative fusion to enhance imagery. Furthermore, it is applicable for LoRA training in the context of image-text to image generation, filling a void in the industry in this area. We further design a simple LightControl to enhance the fidelity of instructional image editing. Finally, extensive experiments demonstrate the effectiveness, efficiency, multifunctionality, and transferability of our X2I. The open-source code and checkpoints for X2I can be found at the following link: https://github.com/OPPO-Mente-Lab/X2I.
\end{abstract}    
\section{Introduction}

\begin{figure*}[ht!]
	\centering
    \includegraphics[width=1.0\textwidth]{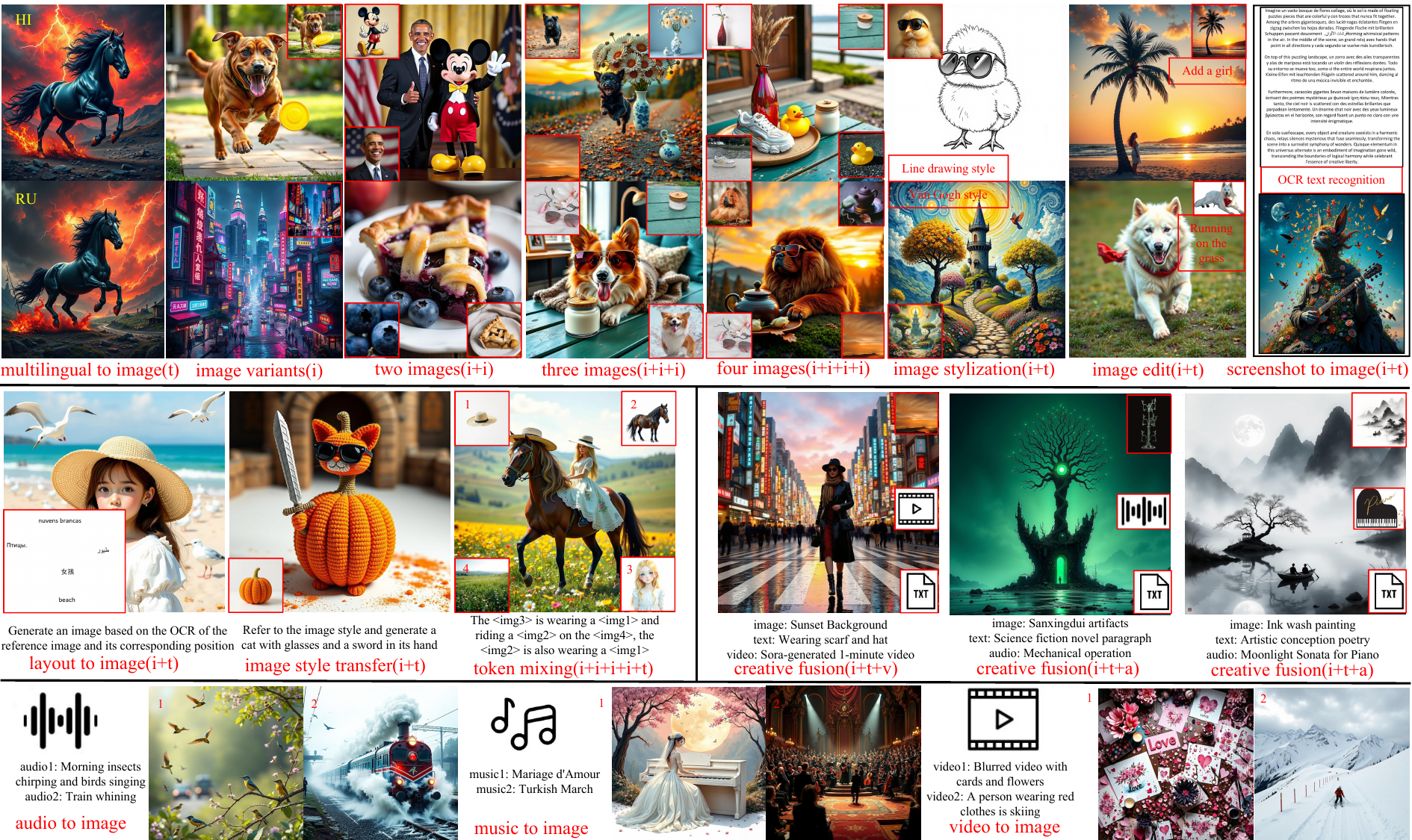}
     \caption{The primary applications of X2I include multilingual text-to-image, image-to-image, image-text-to-image, video-to-image, audio-to-image, and various multimodal combinations for image generation. The red boxes embedded in the images indicate the input images, text, video, or audio. The red box text below the images represents the abbreviated application names, with the initial letters of different modalities in parentheses. The descriptions below some of the images provide a brief overview of the input audio or video.}
  \label{fig:banner}
\end{figure*}

The recently open-sourced T2I models\cite{nichol2021glide, ramesh2022hierarchical, rombach2022high, saharia2022photorealistic,podell2023sdxlimprovinglatentdiffusion}, such as Flux.1\cite{flux2024}, have ushered in a new era of AI art due to their ability to generate realistic, photograde images that are both interesting and creative. The framework of T2I models has evolved from early GAN-based models\cite{karras2019stylebasedgeneratorarchitecturegenerative,esser2021tamingtransformershighresolutionimage}, auto-regressive models\cite{ramesh2021zeroshottexttoimagegeneration,yu2022scalingautoregressivemodelscontentrich}, and UNet-based diffusion models\cite{podell2023sdxlimprovinglatentdiffusion} to DiT models\cite{chen2023pixartalphafasttrainingdiffusion,esser2024scalingrectifiedflowtransformers,kolors}. 
To enhance both controllability and practicality in visual generation, numerous implicit\cite{ma2023glyphdraw,liu2024glyph,ma2025glyphdraw2automaticgenerationcomplex} and explicit\cite{ma2024subject,li2024blip,zhang2023adding,ye2023ip} instruction-based editing models leveraging reference images have been developed quickly. Additionally, several unified image instruction editing models\cite{han2024aceallroundcreatoreditor,shi2024seededitalignimageregeneration,yu2024anyeditmasteringunifiedhighquality} have appeared. They all share a critical characteristic: the requirement of collecting comprehensive instruction-based editing datasets and committing to costly training resources.

The text encoder in T2I is critical for generating semantically relevant images, as seen from early models such as CLIP\cite{radford2021learningtransferablevisualmodels} to later models such as T5\cite{raffel2023exploringlimitstransferlearning} and advanced LLMs\cite{touvron2023llama2openfoundation,deepseekai2024deepseekv2strongeconomicalefficient}. The main goal is to improve the semantic understanding of text without considering the impact of other modalities on visual output. In contrast, VLMs, including early approaches like Flamingo\cite{alayrac2022flamingovisuallanguagemodel} and BLIP2\cite{li2023blip2bootstrappinglanguageimagepretraining}, along with newer models like Llama-Vision\cite{chu2024visionllamaunifiedllamabackbone}, QwenVL\cite{bai2023qwenvlversatilevisionlanguagemodel}, and InternVL\cite{chen2023internvl}, have effectively integrated pre-trained visual encoders with LLMs for enhanced text and visual comprehension. Is there a more streamlined way to transfer VLM or MLLM capabilities to DiT to enable diverse modal inputs? The industry is primarily tackling this challenge through two strategies. One method, similar to PEA-Diffusion\cite{ma2023pea}, utilizes feature distillation by aligning multilingual LLM encoders to the original dimensions using MLP. However, it requires substantial resources of 1600 GPU hours and large-scale image-text datasets. This distillation relies on the U-Net\cite{ronneberger2015u} block's outputs and restricts support for multimodal input. Another method is the VLMs-based replacement of T2I\cite{erwold-2024-qwen2vl-flux}, but it is costly, requiring extensive training or fine-tuning of T2I or MLLMs, necessitating substantial data quality and quantity\cite{shi2024seededitalignimageregeneration,han2024aceallroundcreatoreditor}. Alternatively, a solution akin to GlueGen\cite{qin2023gluegenplugplaymultimodal} could be employed to align the features of a single-modal or multimodal encoder with the latent space of existing T2I models. However, aligning solely through the encoder still results in suboptimal overall performance.

In this work, we introduce an efficient framework, X2I, along with a simple LightControl module to facilitate the enhancement of image transition from weak to strong fidelity. Utilizing a DiT-based image generation model as a teacher, the student model mirrors the entire architecture of the teacher model but substitutes its text encoder with MLLMs. Both models are capable of processing the same linguistic input. To optimize our training data without gathering images, we opt to distill the teacher model's inference abilities. In T2I models, this necessitates performing distillation during the reverse denoising phase. Furthermore, we efficiently transfer the visual generation capabilities of the teacher model to the student model by ingeniously designing the AlignNet structure and distilling the attention. As a result, after alignment, the student model acquires the multimodal comprehension abilities of MLLMs during image generation. Moreover, we formulate a template input format intended for the student model to effectively interpret instructions from different modalities. Since MLLMs generally have proficiency in understanding the global semantic features of visual content, we develop LightControl to derive structured data from reference images. This enhances visual fidelity and facilitates precise and controllable image editing.

After training the X2I, we conduct various experiments to compare the capabilities of T2I and instructional-based image editing. Additionally, we conduct subjective experiments for image generation with various multimodal mixed inputs, including multilingual, image (I2I), image-text (IT2I), video (V2I), and audio (A2I). The results demonstrate that X2I acquires robust multimodal understanding capabilities with less than 1\% performance degradation in the T2I generation.
Furthermore, for more personalized tasks like I2I or IT2I, we conduct experiments involving the training of traditional LoRA\cite{hu2021loralowrankadaptationlarge} to fulfill industry requirements, addressing a training gap in LoRA for these purposes. Moreover, to enhance the adaptability of the framework, only the AlignNet parameters within the model are updated, enabling the aligned student model to accommodate a variety of downstream tasks, such as LoRA, ControlNet\cite{tan2025ominicontrolminimaluniversalcontrol}, IP-Adapter\cite{ye2023ip}, different fine-tuning models, and compression models. Notably, some objective metrics for tasks involving ControlNet and IP-Adapter show improvements over the teacher model. This improvement is due to X2I's robust support for image conditional information input, which facilitates further extraction of image features. We also perform experimental analyses on training efficiency, attaining 98.2\% of the teacher model's performance with 48K training samples.

In summary, our key contributions are as follows.
\begin{itemize}

\item We introduce a novel method for transferring the understanding capabilities of MLLMs to DiT models, achieving fast fitting speeds with only a small amount of monolingual text data through the carefully designed AlignNet and Attention Distillation.
\item We design LightControl to enable the generation of image instruction editing from weak fidelity to high fidelity.
\item The architecture itself possesses strong modularity, leading to performance improvements in certain IT2I tasks. It also has the ability to train LoRA in IT2I tasks.
\item X2I is the first image generation model that supports audio understanding in addition to text and visual understanding.
\end{itemize}

\section{Related Work}

\textbf{Diffusion Models and Text Encoder}. Diffusion models have shown immense potential in the field of T2I. Earlier works\cite{saharia2022photorealistictexttoimagediffusionmodels,rombach2022high,esser2024scalingrectifiedflowtransformers,xue2024raphaeltexttoimagegenerationlarge} use U-Net as the backbone of the diffusion model, which, to some extent, impacts the scalability of the models. Inspired by DiT\cite{peebles2023scalablediffusionmodelstransformers}, many works\cite{chen2023pixartalphafasttrainingdiffusion,flux2024,esser2024scalingrectifiedflowtransformers,li2024hunyuanditpowerfulmultiresolutiondiffusion} replace the U-Net backbone with transformers, greatly improving the quality of the generated images. In order to improve the understanding of the diffusion model's prompts, Stable Diffusion (SD) used CLIP to encode text information, while Flux.1 and eDiff-I\cite{balaji2023ediffitexttoimagediffusionmodels} utilize information encoded by both T5 and CLIP to guide image generation. Recently, works like LuminaT2X\cite{gao2024luminat2xtransformingtextmodality}, Kolors\cite{kolors}, and LI-DiT\cite{ma2024exploringrolelargelanguage} directly use pre-trained decoder-only LLMs as the prompt encoder, which enhances the prompt-following ability in image generation. However, these methods can only encode the text prompt, lacking the ability to understand information from images, videos, audio, etc.

\textbf{MLLMs as Text Encoders for T2I}. VLMs\cite{li2023blip2bootstrappinglanguageimagepretraining,wang2024cogvlmvisualexpertpretrained,bai2023qwenvlversatilevisionlanguagemodel,wang2024qwen2vlenhancingvisionlanguagemodels,liu2023visualinstructiontuning,chen2023internvl,lu2024deepseekvlrealworldvisionlanguageunderstanding,wu2024deepseekvl2mixtureofexpertsvisionlanguagemodels} have made significant progress in visual language understanding tasks. Furthermore, models such as AnyGPT\cite{zhan2024anygptunifiedmultimodalllm}, NextGPT\cite{wu2024nextgptanytoanymultimodalllm}, X-LLM\cite{chen2023xllmbootstrappingadvancedlarge}, and MiniCPM-o\cite{hu2024minicpmunveilingpotentialsmall} have further developed the capabilities in audio understanding. In articles that fuse multimodal models with image generation models, MUMU\cite{berman2024mumubootstrappingmultimodalimage} and ML-MGIE\cite{fu2024guidinginstructionbasedimageediting} inject multimodal information into the image generation model using VLM but also change the weights of the generation model during training, lacking plug-and-play capabilities.
KOSMOS-G\cite{pan2024kosmosggeneratingimagescontext} and Easyref\cite{zong2024easyrefomnigeneralizedgroupimage} achieve alignment while preserving the weights of the original SD\cite{esser2021tamingtransformershighresolutionimage} and SDXL\cite{podell2023sdxlimprovinglatentdiffusion} models. KOSMOS-G trains a VLM and AlignerNet from scratch, whereas Easyref trains the features and projector of the final layer output of the VLM. These methods consume significant computational resources and are limited to a single task, failing to fully leverage the zero-shot capabilities of VLM. Furthermore, the aforementioned methods focus on alignment with U-Net-based models like SDXL. Currently, there is a lack of an efficient alignment method for DiT models, such as Flux.1 and SD3. Work in the related field, such as HunyuanVideo\cite{kong2025hunyuanvideosystematicframeworklarge}, requires an additional CLIP text encoder for alignment. Qwen2VL-Flux\cite{erwold-2024-qwen2vl-flux} not only retains the structure of T5 but also alters the weights of Flux.1 itself, significantly reducing the model's plug-and-play and zero-shot capabilities.

\textbf{Distillation of DiT}. Knowledge distillation (KD) has been widely applied in diffusion models\cite{zheng2024trajectoryconsistencydistillationimproved,kim2024consistencytrajectorymodelslearning,zhu2024slimflowtrainingsmalleronestep}.
SSKD\cite{wang2022attentiondistillationselfsupervisedvision} involves the direct distillation of critical attention mechanism information from teacher to student, which can considerably reduce the performance gap between both.
EFFICIENT-VDIT\cite{ding2025efficientvditefficientvideodiffusion} utilizes a multi-step consistency distillation technique to accelerate DiT sampling. PEA-Diffusion integrates multilingual CLIP with SDXL by employing the L2-norm distance, thus allowing the T2I model to support multilingual capabilities. However, this method is specifically tailored for the U-Net backbone and demands significant computational resources.

\section{Methodology}

\subsection{Preliminary}


The diffusion process is performed in the latent space, where a transformer denoiser $\epsilon_\theta$ is employed to predict noise $\epsilon$ with the current timestep $t$, noisy latent $x$ and generation conditions, $c$ and $c_p$, where $c_p=\tau_\theta(y)$ is produced by encoding the text prompts $y$ with a pre-trained CLIP text encoder $\tau_\theta$, and $c=\varepsilon_\theta(y)$ is produced by encoding the text prompts $y$ with a pre-trained T5 text encoder $\varepsilon_\theta$.
To enhance the capability of feature fusion, MM-DiT\cite{esser2024scalingrectifiedflowtransformers} uses adaptive layer normalization (AdaLN) to improve the adaptability of the model when processing different input conditions. There are two fully connected layers plus a SiLU activation function forming the feature extraction and normalization networks $\zeta_\theta$ and $\delta_\theta$ , which respectively extract features for $x$ and $c$. These features are used as the normalization and scaling parameters for the multi-head self-attention, as shown below:
\begin{equation}
\begin{aligned}
x, \alpha_1, \beta_2, \gamma_2, \alpha_2 &= \zeta_\theta(x,t,c_p) \\
c, \alpha_{1c}, \beta_{2c}, \gamma_{2c}, \alpha_{2c} &= \delta_\theta(c,t,c_p),
\end{aligned}
\label{eq:1}
\end{equation} then, concatenate $x$ and $c$ directly with $\varphi(\cdot)$ and use the self-attention to achieve feature interaction and fusion. The output is then split according to the corresponding indices, as shown below:
\begin{equation}
\begin{aligned}
    x_{A},c_{A} = {softmax}\left(\frac{QK^T}{\sqrt{d_0}}\right)\cdot V,
\end{aligned}
\label{eq:2}
\end{equation}
where $Q=W_Q\cdot\varphi(x,c), K=W_K\cdot\varphi(x,c), V=W_V\cdot\varphi(x,c)$, $x_{A},c_{A}$ represents the feature value output on the image and text sides obtained after computing attention and truncating according to the respective index values. $W_Q, W_K, W_V$ are learnable projection matrices. 

Both the text and image sides undergo the following process, and only the image-side processing is described below for the convenience of presentation. After the residual module, the regressed scaling parameter $\alpha_1$ is used to scale the weights of the image-side input,
\begin{equation}
\begin{aligned}
    x_{LN} = \mu(x + \alpha_1*x_{A}),
\end{aligned}
\label{eq:3}
\end{equation}
where $\mu$ represents Layer Normalization (LN). Then, normalization is performed utilizing adaptive parameters $\beta_2,\gamma_2$,
\begin{equation}
\begin{aligned}
    x_{FF} = \nu(x_{LN}*(1+\gamma_2)+\beta_2),
\end{aligned}
\label{eq:4}
\end{equation}
where $\nu$ represents FeedForward 
 (FF). Finally, the output of the current block is obtained by multiplying the regression scaling coefficients and the residuals
\begin{equation}
\begin{aligned}
    x_{O} = (x + \alpha_1*x_{A}) + x_{FF}*\alpha_2. 
\end{aligned}
\label{eq:5}
\end{equation}

\subsection{Model Overview}
\begin{figure}[ht!]
	\centering
    \includegraphics[width=0.5\textwidth]{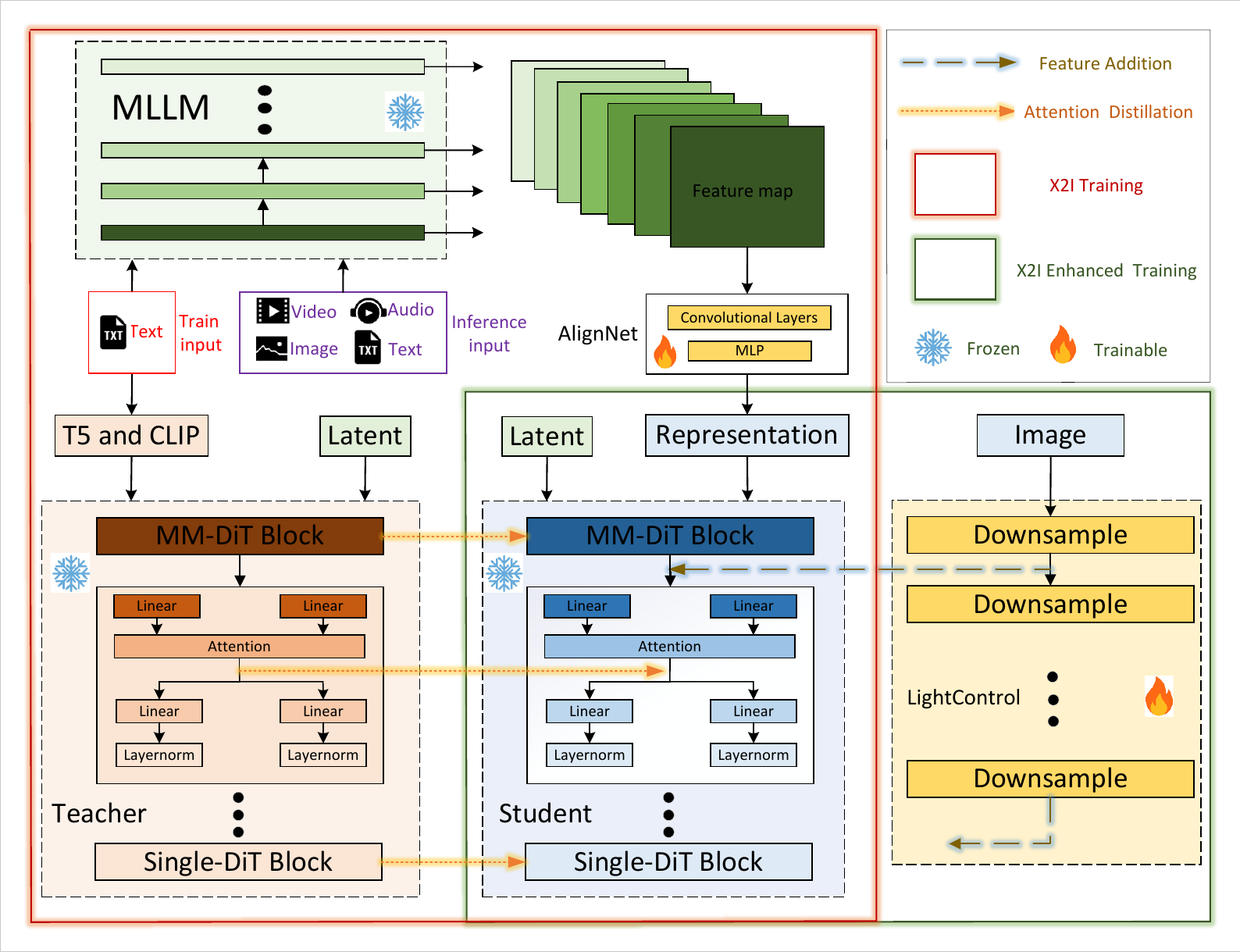}
     \caption{Overview diagram of the X2I. In the deep red box on the left is the X2I training, where only the text data needs to be input for training, with trainable parameters limited to AlignNet and the distillation location situated within the attention output of MM-DiT. The light green box at the bottom right represents X2I enhanced training. After the X2I training, AlignNet is fixed, and the trainable parameters are limited to LightControl.}
  \label{fig:mllm_flux_new}
\end{figure}

The framework is composed of two parts, as shown in \cref{fig:mllm_flux_new}. The initial part involves X2I training on a textual dataset, with AlignNet serving as trainable parameters. Distribution alignment between the student and teacher models is achieved through attention distillation. The subsequent part incorporates the LightControl module, forming X2I Enhanced Training, which enhances the precision of reference images in tasks involving image instruction editing.

\subsection{AlignNet with MLLM's Hidden State}
Many approaches use MLLM as text encoders for T2I or feature extractors for other modules, typically utilizing only the last\cite{pan2024kosmosggeneratingimagescontext} or penultimate\cite{xia2024dreamomniunifiedimagegeneration} hidden states. However, according to Oscar Skean's analysis\cite{skean2024doesrepresentationmatterexploring}, the quality of hidden states in the intermediate layers of transformer-based LLMs is often superior to that of the final and initial layers. Recognizing that employing a static intermediate layer's representation might be suboptimal for adapting to diverse MLLM models, we propose utilizing representations from every layer of MLLMs as input to AlignNet.

To integrate hidden states from various layers of MLLMs, we design a simple CNN to capture spatial and channel-wise interactions across layers. Convolutional operations allow the extraction of intricate patterns and enhance the rationale of weight distribution, permitting spatial flexibility and forming inter-layer linkages. By adjusting the kernel size, we can also generate cross-layer receptive field connections, thus attaining superior multi-scale integration.

In MLLM, define a feature extractor that projects text $y$ into a middle feature $H \in \mathbb{H}^{b \times m \times s \times z}$, where $m$ represents all layers of MLLM's hidden states, each layer has a hidden size of $z$ and a sequence length of $s$. Next, we define a CNN mapping network $\Psi$ to enable self-learning of weights across all layers' hidden states, resulting in the aggregation of hidden states from $m$ layers into $1$ layer of weighted hidden states as the following formula,
\begin{equation}
\begin{aligned}
    y_p,y = \Phi(\Psi (m,1,k,p)(H)),
\end{aligned}
\label{eq:7}
\end{equation}
where the kernel size is $k$, and the padding is denoted as $p$, $\Phi$ is a simple MLP network that maps the feature learned by the $\Psi$ with weight information to the corresponding dimension of $c$ and ${c_p}$, thereby obtaining new features $y$ and ${y_p}$.

\subsection{Attention Distillation}
KD can be divided into two main categories: logits distillation and feature distillation. 
In MM-DiT, intermediate layer features encompass low-level information such as pixel values, shape, and gradient information, as well as high-level information such as color, theme, and lighting details. In this paper, the student model is conditioned to inject multimodal feature information. Knowledge transfer occurs layer by layer through the concatenation of the information injected at each level. Therefore, for deeply encoded information that logits cannot express, we opt for intermediate feature distillation between layers to achieve alignment.

Further, each MM-DiT consists of AdaLN as \cref{eq:1}, self-attention layer as \cref{eq:2}, LN as \cref{eq:3}, and FF layers as \cref{eq:4}. In addition to regressing the $\gamma$ and $\beta$ after the LN, AdaLN also regresses an $\alpha$ before the end of each residual module. Since biases here directly affect the distribution of features, we choose to perform feature distillation in the position of the attention output $x_A,c_A$ in order to minimize unnecessary regression and maintain consistency in the distribution affected by normalization parameters. This provides an important conclusion in exploring direct feature distillation in the MM-DiT feature layers as well as the the single-DiT block of Flux.1.

Regarding the choice of attention distillation loss, we adopt reverse KL (RKL) from MiniLLM\cite{gu2024minillmknowledgedistillationlarge}. KL performs well in traditional tasks due to the smaller output space and fewer modes in conventional classification tasks. However, for LLMs, the output space is more complex with more modalities; RKL can prevent the student model from overfitting to the low-probability outputs of the teacher model, and the same applies to the MM-DiT structure. For the Flux.1 structure, the target loss for $l$ layers of MM-DiT is defined as follows:
\begin{equation}
\mathcal{L}(\theta) = \text{KL}[\eta_l || \upsilon_l]
= \left[ - \mathbb{E} \log \frac{\upsilon_l(A_t|\epsilon,c)}{\eta_l(A_s|\epsilon,y)} \right],
\end{equation}
where $\eta,\upsilon$ specifically represent the attention output mappings in each layer of the student and teacher models in MM-DiT, respectively. $A_t$ denotes the attention output $x_A,c_A$ mentioned above; similarly, $A_s$ corresponds to the output of the attention output in the student model.
$\epsilon$ represents noise input because we perform distillation during the inference process, which is the reverse denoising process.

\subsection{Training Stages}

For most pre-trained MLLMs, features from different modalities are mapped to a common feature space, where similar information from different modalities is brought closer together. This implies that different modality features share a common semantic space. Based on this consensus, we opt to perform alignment by inputting pure text modality while simultaneously aligning the inferential capabilities of the teacher model, that is, aligning capabilities during the process of pure noise denoising. The Appendix \cref{sec:feature} \cref{fig:feature} provides a more detailed analysis of the changes in the distribution of different modality spaces before and after alignment, which effectively explains how X2I gained the multimodal understanding capabilities of MLLM simply through textual semantic alignment.

Since MLLMs tend to understand images more from a global semantic perspective, they often lack the crucial fine-grained understanding needed for image generation. To address this issue, we design a simple LightControl in parallel with the MM-DiT structure named X2I enhanced training. This pathway is a convolutional module composed of ResNet blocks, similar to those mentioned in ControlNeXt\cite{peng2024controlnextpowerfulefficientcontrol}. We initialize a ResNet network with 19 layers, mirroring those used in MM-DiT. We define the ResNet module $\Gamma$ to extract the conditional controls; similar to ControlNet, the output of each ResNet module is:
\begin{equation}
y_c^l = \Gamma^l (c_p, c_i; \Theta),
\end{equation}
where $y_c^l$ is the feature output of $l$ the layer of the ResNet, $c_p$ is the corresponding conditional text input, $c_i$ is the conditional reference image input, and $\Theta$ is the corresponding trainable parameter. The final output here is also similar to ControlNet, where $y_c$ and $x_O$ are added feature-wise.

\section{Experiments}

\subsection{Datasets and Implementation}
We randomly select 100K image-text pairs from the Laion2B 
\cite{schuhmann2022laion5bopenlargescaledataset}, and utilize InternVL to generate new captions from the images for training X2I. Furthermore, based on the T2I-Adapter's\cite{mou2023t2iadapterlearningadaptersdig} sketch-guided model and SDXL, we construct a training dataset of 400k images containing five styles: Monet, Baroque, Cartoon, Pixar, and Van Gogh. For the text encoder in the student model, we utilize open-source models from the InternVL series, Qwen2.5-VL-7B\cite{bai2025qwen25vltechnicalreport}, and MiniCPM-o, for the DiT structure, we use Flux.1. To optimize performance, inference ability is distilled by focusing on the denoising process between steps 0 and 1. The X2I is trained on 8 A100 GPUs for 24K steps. X2I training requires over 80GB of GPU memory. We optimize this by separating training and inference, and overlapping communication, achieving a global batch size of 6. Acceleration techniques are detailed in the appendix \cref{sec:Acceleration}.

\subsection{Tasks and Benchmarks}
\label{sec:Metrics}

\textbf{T2I and Multilingual T2I}. \textbf{Evaluation Datasets}. We randomly select five sub-evaluation datasets from T2I-CompBench++\cite{10847875}, each containing 1k images. These include: complex compositions bench, non-spatial bench, attributes binding bench related to color and texture categories, numeracy bench. GenAI-Bench\cite{li2024genaibenchevaluatingimprovingcompositional}, EvalMuse\cite{han2024evalmuse40kreliablefinegrainedbenchmark}, and Multilingual-General\cite{ma2023pea}.
\textbf{Metrics} includes four main categories: image-text matching score indicators like ClipScore (CS) with ViT-bigG\cite{cherti2022reproducible} and FGA-BLIP2 (FB)\cite{han2024evalmuse40kreliablefinegrainedbenchmark} Score.
Human-consistency feedback score indicators include PickScore (PS)\cite{kirstain2023pickapicopendatasetuser}, HPSv2\cite{wu2023humanpreferencescorev2}, and ImageReward (IR)\cite{xu2023imagerewardlearningevaluatinghuman}.
Fine-grained score indicators have BLIP-VQA (BV)\cite{10847875} and UniDet\cite{10847875}.
Model scoring indicators reflecting complex textual semantic alignment such as GPT-4o and VQAScore (VS)\cite{lin2024evaluatingtexttovisualgenerationimagetotext}, with GPT-4o instructions specified by the T2I-CompBench++ for complex compositions evaluation. Furthermore, the IR score ranges from 98.2\% probability within [-2, 2], and the FB score ranges from [1, 5]. We use eight normalized metrics to determine the relative gap compared to the teacher model, defining a Performance Ratio (PR) metric as a performance percentage indicator. 
\textbf{Baseline model} for universal T2I capability comparison is the Flux.1 teacher model, used to assess reduction in student model's basic T2I capabilities. Qwen2VL-Flux has similar functions, but MLLM does not support text input and relies solely on the source model's T5. Thus, it is excluded from comparison. 
For the comparison of multilingual T2I capabilities, we evaluate both PEA-Diffusion and Sana\cite{xie2024sanaefficienthighresolutionimage}.

\textbf{IT2I and I2I}. 
\label{sec:IT2I_main} \textbf{Evaluation Datasets}. We chose DreamBench\cite{ruiz2023dreamboothfinetuningtexttoimage} and we generate six images for each text prompt\cite{li2024blip}.
\textbf{Metrics}. We report the average DINO, CLIP-I, and CLIP-T scores based on all pairs of real and generated images. 
\textbf{Baseline model}. The comparison model includes methods based on different frameworks such as SD, Imagen, Phi-3, SD3, and Flux.1.
In the appendix \cref{sec:IT2I}, we also compare the differences in Qwen2VL-Flux's capabilities in image understanding and image editing based on objective metrics.

\textbf{Image stylization}. \textbf{Evaluation Datasets} consists of 50 images from MSCOCO\cite{lin2015microsoftcococommonobjects} and 50 photorealistic images generated by DiffArtist\cite{jiang2024diffartistaestheticaligneddiffusionmodel}. \textbf{Baseline model} is also selected DiffArtist.

\textbf{ControlNet}. \textbf{Evaluation Datasets} is sourced from MultiGen-20M\cite{qin2023unicontrolunifieddiffusionmodel} and includes four tasks:
A model based on canny edge detection, evaluated using the F1 score. Models based on hed, evaluated using Structure Similarity Index Measure (SSIM). A model based on depth information is evaluated using RMSE.


\subsection{Quantitative Results}
\label{sec:Quantitative}

\textbf{Universal T2I Generation}. As shown in \cref{tab:Universal}, the one that performs closest to the teacher model is the complex compositions evaluation set. Additionally, on some sub-evaluation sets and certain metrics, the performance of X2I even surpass that of the Flux.1. The overall average PR across the five evaluation sets is 99.21\% of that of Flux.1, with such subtle differences being virtually indistinguishable subjectively. Appendix \cref{sec:GenAI} \cref{tab:Universal2} displays the results from the GenAI-Bench and EvalMuse evaluation sets, with performance scores also exceeding 99\%. This further demonstrates that X2I's results in T2I generation are nearly indistinguishable from those of the teacher model Flux.1.

\begin{table*}[ht!]
\centering
\resizebox{0.8\linewidth}{!}{
\begin{tabular}{cccccccccccc}
\hline
\begin{tabular}[c]{@{}c@{}}T2I++\\ CompBench\end{tabular} & Methods  & CS     & PS     & HPSv2  & IR     & FB     & GPT-4o & VS     & BV     & UniDet & PR                        \\ \hline
\multirow{2}{*}{Complex}                                  & Flux.1 & 0.4279 & 0.2297 & 0.2055 & 0.8354 & 3.6990  & 0.9181 & 0.8799 & 0.5402 & -      & \multirow{2}{*}{99.77\%} \\
                                                          & X2I    & 0.4234 & 0.2283 & \textbf{0.2096} & 0.8124 & 3.6723 & 0.9142 & 0.8779 & \textbf{0.5417} & -      &                          \\ \hline
\multirow{2}{*}{Non-spatial}                              & Flux.1 & 0.4418 & 0.2751 & 0.2472 & 1.3013 & 3.5963 & 0.8962 & 0.9075 & 0.6762 & -      & \multirow{2}{*}{99.12\%} \\
                                                          & X2I    & 0.4363 & 0.2706 & 0.2462 & 1.2786 & 3.5449 & 0.8821 & 0.8996 & 0.6571 & -      &                          \\ \hline
\multirow{2}{*}{Color}                                    & Flux.1 & 0.4733 & 0.2872 & 0.2417 & 1.4179 & 4.2268 & 0.7060 & 0.9077 & 0.6642 & -      & \multirow{2}{*}{99.72\%} \\
                                                          & X2I    & 0.4691 & \textbf{0.2881} & \textbf{0.2421} & 1.3872 & 4.1845 & \textbf{0.7113} & 0.9014 & \textbf{0.6643} & -      &                          \\ \hline
\multirow{2}{*}{Texture}                                  & Flux.1 & 0.4572 & 0.2589 & 0.2411 & 1.1994 & 4.0601 & 0.7588 & 0.8807 & 0.5281 & -      & \multirow{2}{*}{99.18\%} \\
                                                          & X2I    & 0.4504 & 0.2570 & 0.2401 & 1.1287 & 4.0055 & 0.7447 & \textbf{0.8809} & 0.5172 & -      &                          \\ \hline
\multirow{2}{*}{Numeracy}                                 & Flux.1 & 0.4730 & 0.2664 & 0.2431 & 1.3459 & 3.4852 & 0.7860 & 0.7841 & 0.4528 & 0.6112 & \multirow{2}{*}{99.36\%} \\
                                                          & X2I    & 0.4713 & \textbf{0.2670} & \textbf{0.2485} & 1.2723 & 3.4140 & 0.7680 & 0.7798 & 0.4464 & \textbf{0.6206} &                          \\ \hline
\end{tabular}}
\caption{Objective performance on the T2I-CompBench++ evaluation dataset.}
\label{tab:Universal}
\end{table*}

\textbf{Multilingual T2I Generation}
\cref{tab:Multilingual} includes three multilingual evaluation datasets from Multilingual-General, as well as four additional translations in German, Portuguese, Spanish, and French. While the performance in English remains consistent with a one percentage point difference similar to before. For more details, please refer to \cref{sec:Multilingual}.

\textbf{IT2I Generation}
\cref{tab:DreamBench} shows the performance of different methods on the DreamBench. Overall, X2I has a slight advantage in DINO scores. CLIP-I shows a two-point difference compared to KOSMOS-G and there is a considerable gap in CLIP-T compared to recent methods like OmniControl and IP-Adapter based on Flux.1, indicating that X2I still has room for improvement in its ability to faithfully follow the semantics of images.

\begin{table}[ht!]
\centering
\resizebox{0.9\linewidth}{!}{
\begin{tabular}{ccccc}
\hline
Methods           & Base                    & DINO           & CLIP-I         & CLIP-T         \\ \hline
Textual Inversion\cite{yang2023controllabletextualinversionpersonalized} & \multirow{7}{*}{SD\cite{rombach2022high}}     & 0.569          & 0.780          & 0.255          \\
DreamBooth\cite{ruiz2023dreamboothfinetuningtexttoimage}        &                         & 0.668          & 0.803          & 0.305          \\
Custom Diffusion\cite{kumari2023multiconceptcustomizationtexttoimagediffusion}  &                         & 0.643          & 0.790          & 0.305          \\
BLIP-Diffusion\cite{li2024blip}    &                         & 0.594          & 0.779          & 0.300          \\
Subject-Diffusion\cite{ma2024subject} &                         & 0.711          & 0.787          & 0.293          \\
IP-Adapter\cite{ye2023ip}        &                         & 0.667          & 0.813          & 0.289          \\
KOSMOS-G\cite{pan2024kosmosggeneratingimagescontext}          &                         & 0.694          & \textbf{0.847} & 0.287          \\ \hline
SuTI\cite{chen2023subjectdriventexttoimagegenerationapprenticeship}              & Imagen\cite{saharia2022photorealistictexttoimagediffusionmodels}                  & 0.741          & 0.819          & 0.304          \\ \hline
OmniGen\cite{pan2024kosmosggeneratingimagescontext}           & Phi-3\cite{abdin2024phi3technicalreporthighly}                   & -              & 0.801          & 0.315          \\ \hline
UNIC-Adapter\cite{duan2024unicadapterunifiedimageinstructionadapter}      & SD3\cite{esser2024scalingrectifiedflowtransformers}                     & 0.816          & 0.841          & 0.306          \\ \hline
IP-Adapter\cite{ye2023ip}        & \multirow{3}{*}{Flux.1\cite{flux2024}} & 0.768          & 0.803          & 0.322          \\
OminiControl\cite{tan2025ominicontrolminimaluniversalcontrol}      &                         & 0.740          & 0.768          & \textbf{0.329} \\
X2I               &                         & \textbf{0.817} & 0.826          & 0.304          \\ \hline
\end{tabular}}
\caption{Performance of different methods on DreamBench.}
\label{tab:DreamBench}
\end{table}

\subsection{Qualitative Results}
\label{sec:Qualitative}
\textbf{X2I} training achieves outstanding results through feeding pure text. \cref{fig:banner} illustrates the effects of some of these functions. T2I possesses the capability to support over 29 natural languages, covering the major language families worldwide; I2I includes image variations and the fusion of concepts from multiple images. IT2I encompasses single image instruction editing, style transfer, image stylization, instruction editing for multiple images, and the result of text token mixing with multiple images. It integrates multiple visual concepts to create unique images, extracts specific concepts from a set of input images, and synthesizes these concepts into new images, allowing precise control over the visual details of the images and avoiding common vagueness and uncertainty in text descriptions. Examples include changing clothes in the combination of clothes and people, performing group photos in person combinations, and creative conceptual combinations with three or more images.

A2I enables auditory information to be visualized, providing a way to ``see sound". This aspect is not limited to regular audio information but also processing various audio types, including natural water flow sounds, animal calls, iconic IP character voices, and the emotions that different music pieces convey, translating these into corresponding visual images. V2I can inductively summarize single frames from videos as images and generate high-definition images from low-resolution videos.

Additionally, more creative combinations can be achieved. In the middle of \cref{fig:banner}, the last three examples in the second row illustrate this: the first input example is a 1-minute video generated by Sora\cite{liu2024sorareviewbackgroundtechnology}, a sunset background image along with the text instruction ``wearing a hat and scarf", the generated image successfully maintains the original video's semantic information while adhering well to both the text and image instructions. The second input combines an image of the Sanxingdui bronze sacred tree, a sci-fi novel text fragment, and a sound of mechanical operation, resulting in an output that blends all the semantic information, adding a mysterious touch. The third input is a Chinese ink painting, a classical poem, and Beethoven's Moonlight Sonata, resulting in a final image that perfectly integrates the three modalities into an evocative artwork. For more interesting effect displays, please refer to the appendix \cref{sec:zero}.

\textbf{X2I enhanced Training}. The purpose of constructing LightControl is to enhance the fidelity of image instruction editing. Essentially, it is not limited to a specific downstream task. In this paper, relevant experimental validation is only conducted in the stylization domain.  For more comparative results, please refer to the appendix \cref{sec:style}.


\textbf{IT2I LoRA}. For highly personalized IT2I image generation tasks tailored to the user, we can leverage the X2I to train a LoRA. In this paper, we train X2I using reference images in the style of abstract art line drawings. More comparative results, please refer to the appendix \cref{sec:LoRA}.

\textbf{Plug-and-Play}. In this paper, we have validated the prominent variant models and related downstream models in the Flux.1 community. Specifically, we compare objective metrics for tasks involving ControlNet and IP-Adapter. Additionally, incorporating the original image as input in the X2I task resulted in improvements in both objective and subjective metrics. More results in the appendix \cref{sec:Downstream}.

\textbf{Reasoning ability and multi-turn dialogue generation ability}. MLLMs themselves possess certain reasoning abilities and multi-turn dialogue capabilities. X2I also inherits some of these related abilities.  More results in the appendix \cref{sec:reasoning}.

\subsection{X2I Training Convergence}
\begin{figure}[ht!]
	\centering
\resizebox{1\linewidth}{!}{
    \includegraphics[width=0.5\textwidth]{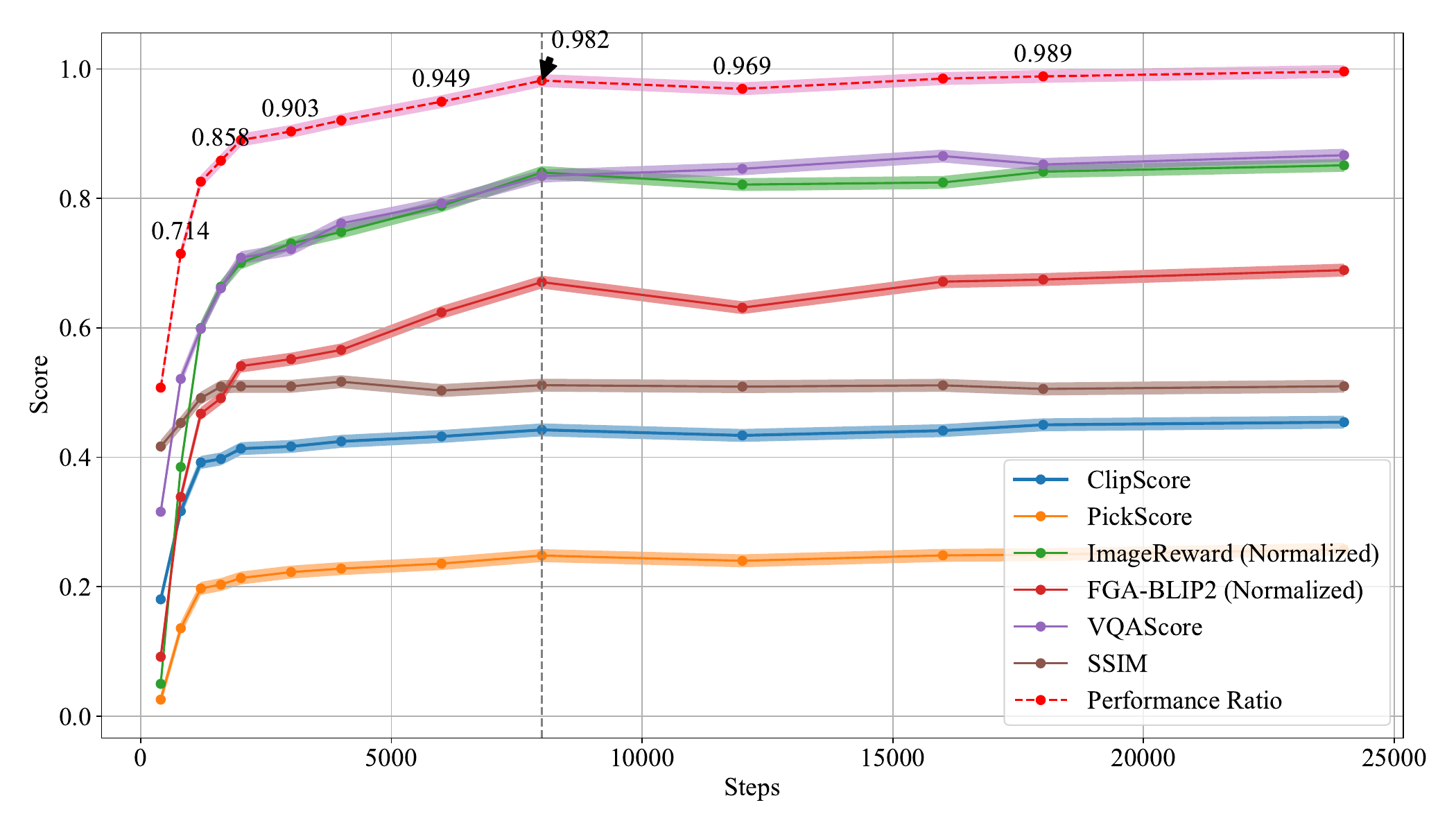}}
     \caption{X2I training performance vs. training steps.}
  \label{fig:Convergence}
\end{figure}

Benefitting from the design of AlignNet and attention distillation, the training convergence speed of X2I is remarkably fast. Here, we have calculated the performance metrics including CS, FB Score, PS, IR, and VS, and used the average of these five objective metrics as the Performance Ratio. Additionally, we have also calculated the variation of the SSIM metric. As shown in \cref{fig:Convergence} with Qwen2.5-VL-7B, the convergence speed in the early stages of training exhibits an almost linear growth, achieving a Performance Ratio of 98.2\% with only 8K steps of training. Appendix \cref{sec:internVL} further demonstrates the experimental results of InternVL\cite{chen2023internvl} with different capacity adaptations for X2I.




\subsection{Ablation Study}

\begin{table}[ht!]
\resizebox{1\linewidth}{!}{
\begin{tabular}{ccccccc}
\hline
Models            & CS              & PS              & IR              & FB              & SSIM            & PR                \\ \hline
FLUX.1            & 0.4517          & 0.2578          & 1.3891          & 3.7789          & -               & 100\%            \\ \hline
A1                & 0.4313          & 0.2339          & 1.1149          & 3.4946          & 0.5089          & 95.40\%          \\
A3                & 0.4348          & 0.2407          & 1.2198          & 3.544           & 0.5131          & 96.62\%          \\
A3\_ada         & 0.4429          & 0.2447          & 1.2278          & 3.6055          & 0.5131          & 97.36\%          \\
A3\_t5\_ada     & 0.4397          & 0.2452          & 1.2757          & 3.6016          & 0.5036          & 97.57\%          \\
A29\_ada        & 0.4463          & 0.2478          & 1.2954          & 3.6789          & 0.5271          & 98.40\%          \\
A29\_t5\_ada    & 0.4361          & 0.2421          & 1.1988          & 3.5712          & 0.5084          & 96.73\%          \\ 
\textbf{A29\_CNN} & \textbf{0.4508} & \textbf{0.2518} & \textbf{1.3799} & \textbf{3.6748} & \textbf{0.5189} & \textbf{99.11\%} \\ \hline
Block             & 0.1483          & -0.0220         & -2.2800         & 1.0854          & 0.0475          & 45.65\%          \\
FF                & 0.2995          & 0.1027          & -0.7019         & 2.2326          & 0.4623          & 69.58\%          \\
LN              & 0.4392          & 0.2495          & 1.3004          & 3.6557          & 0.5128          & 98.16\%          \\
Oneside           & 0.4470          & 0.2461          & 1.3030          & 3.6312          & 0.5077          & 98.13\%          \\ \hline
KL                & 0.4528          & 0.2532          & 1.4039          & 3.6988          & 0.5152          & 99.26\%          \\
JS                & 0.4490          & 0.2537          & 1.3795          & 3.6972          & 0.5202          & 99.26\%          \\ 
\textbf{RKL}      & \textbf{0.4543} & \textbf{0.2577} & \textbf{1.4043} & \textbf{3.7568} & \textbf{0.5208} & \textbf{99.70\%} \\ \hline
\end{tabular}}
\caption{Three ablation results of X2I.}
\label{tab:Ablation}
\end{table}

In this paper, we design three types of ablation experiments: AlignNet structure with the number of feature layers extracted by MLLMs, the feature distillation position in the MM-DiT structure, and different target losses for feature distillation. The evaluation dataset uses the Multilingual-General. The evaluation metrics are CS, FB Score, PS and IR, and the SSIM metric for image variants. Additionally, the overall PR of these four normalized metrics. The MLLMs chosen is Qwen2.5-VL.

\textbf{AlignNet Structure Ablation}.
\cref{tab:Ablation}  upper portion show AlignNet structure ablation. ``A1" indicates extracting only the last layer features of Qwen2.5-VL while the AlignNet structure consists of a simple MLP, similar to PEA-Diffusion, performing only simple dimensional mapping. The overall performance of the relevant metrics for T2I generation is 95.4\% of the teacher, showing a relatively large decline. ``A3" extracts features from the first and the last two layers of Qwen2.5-VL, where the first layer features are static token-mapped vector features trained by MLLMs, containing the most original token information, but lacking contextual information. Therefore, combining with the last two layers features offers complementary advantages. By directly taking the average pooling of the features, the performance improvement is about one percentage point compared to ``A1". ``A3\_ada" further learns adaptive weights for each layer of features, improving by 0.74\%. ``A3\_t5\_ada" increases the nonlinearity complexity of AlignNet by adding a custom T5 module for feature mapping, but the improvement is minimal. ``A29\_ada" selects features from all 29 layers of Qwen2.5-VL, leading to about a one percentage point improvement over ``A3\_ada". However, ``A29\_t5\_ada" shows some metric decline, suggesting that deep nonlinear complex mapping is not required for such modal alignment learning. Furthermore, the scalar parameter learning fails to capture the synergy between low-level edge features and high-level semantic features, as well as issues of spatial insensitivity and poor dynamic adaptability. Therefore, ``A29\_CNN" ultimately adopts a CNN to further enhance the non-linear learning of each layer's feature weights, resulting in optimal performance, reducing the performance loss compared to the teacher to less than one percentage point.

\textbf{Distillation Position Ablation}.
The middle four lines of \cref{tab:Ablation} show distillation position ablation. ``Block" refers to direct distillation of each layer's MM-DiT and Single-DiT output positions in Flux.1, analogous to PEA-Diffusion’s method of distilling each U-Net block output in SDXL. The experimental result shows a suboptimal performance of 45.65\%. ``FF" denotes distillation at the output position of each block's FeedForward layer, as shown in \cref{eq:4}, results indicate a significant improvement compared to ``Block" distillation. This is likely due to the influence of the regression scaling factor $\alpha_2$ on distribution alignment in \cref{eq:5}. LN denotes distillation at the second layer norm output position of each block, i.e., the output of \cref{eq:3}, approaching optimal results but still possibly affected by the regression scaling factor $\alpha_2$. ``Oneside" indicates distillation solely at the conditional side output after MM-DiT's self-attention, aiming to explore whether the interaction information’s purest form can be obtained after self-attention and token length segmentation. Experimental results remain about one percentage point below the optimal results.

\textbf{Feature Alignment Loss Ablation}.
We additionally compare three other target loss functions: KL, JS, and RKL divergence. Both KL divergence and JS divergence consistently exhibited slightly better performance compared to the MSE target loss like ``A29\_CNN" before. However, the best performance was achieved with RKL, which outperformed KL divergence by 0.44\% percentage points.

\section{Conclusion and Limitations}
In this paper, we introduce a novel framework X2I for aligning MLLMs and the DiT visual generation model. This framework can be fully trained with minimal training resources and minimal training data. Our extensive experimental results demonstrate that X2I maintains its general performance while extending additional functionalities, including multilingual image generation, image instruction editing, conceptual fusion of multiple images, image generation from videos or audio, and further exploration of creative combinations from different modalities to unleash imagery with X2I. Additionally, X2I supports LoRA training for multiple modalities, enabling seamless integration with various controllable downstream tasks in the open-source community. Furthermore, we have developed LightControl to explore more precise and controllable possibilities for visual generation.

The limitations of this paper lie in the somewhat inferior precision and control of X2I in the unified direction of image instruction editing, as well as the failure to fully explore the logical reasoning, few-shot learning, and chain-of-thought capabilities of MLLMs. In the future, we will continue to delve deeper into this direction.
\clearpage 

{
    \small
    \bibliographystyle{ieeenat_fullname}
    \bibliography{main}
}

\clearpage
\setcounter{page}{1}
\maketitlesupplementary

\section{X2I's reasoning ability and multi-turn dialogue generation ability}
\label{sec:reasoning}

\begin{figure}[ht!]
	\centering
\resizebox{1\linewidth}{!}{
    \includegraphics[width=1\textwidth]{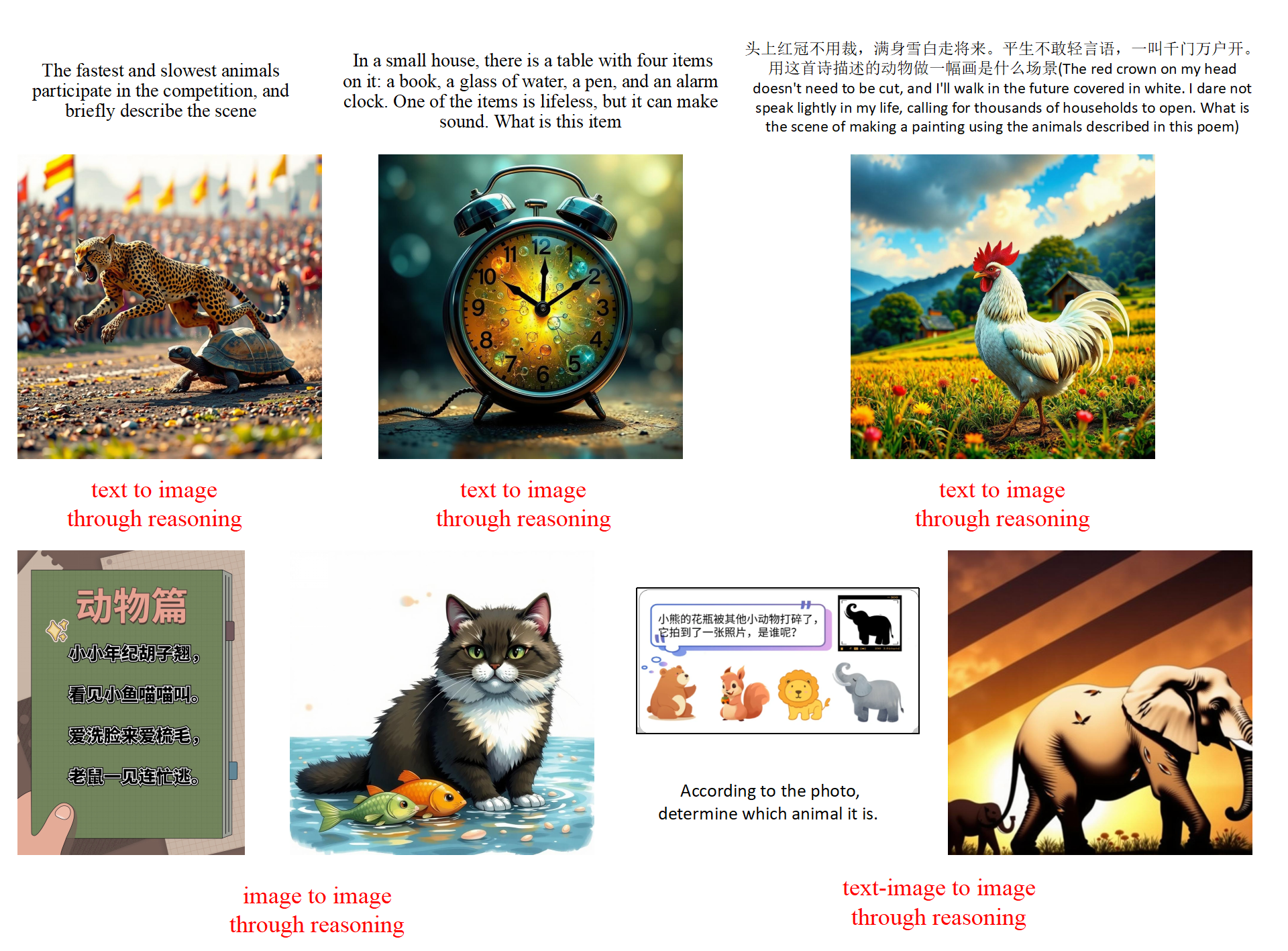}}
     \caption{X2I's reasoning ability.}
  \label{fig:reasoning}
\end{figure}

\begin{figure}[ht!]
	\centering
\resizebox{1\linewidth}{!}{
    \includegraphics[width=1\textwidth]{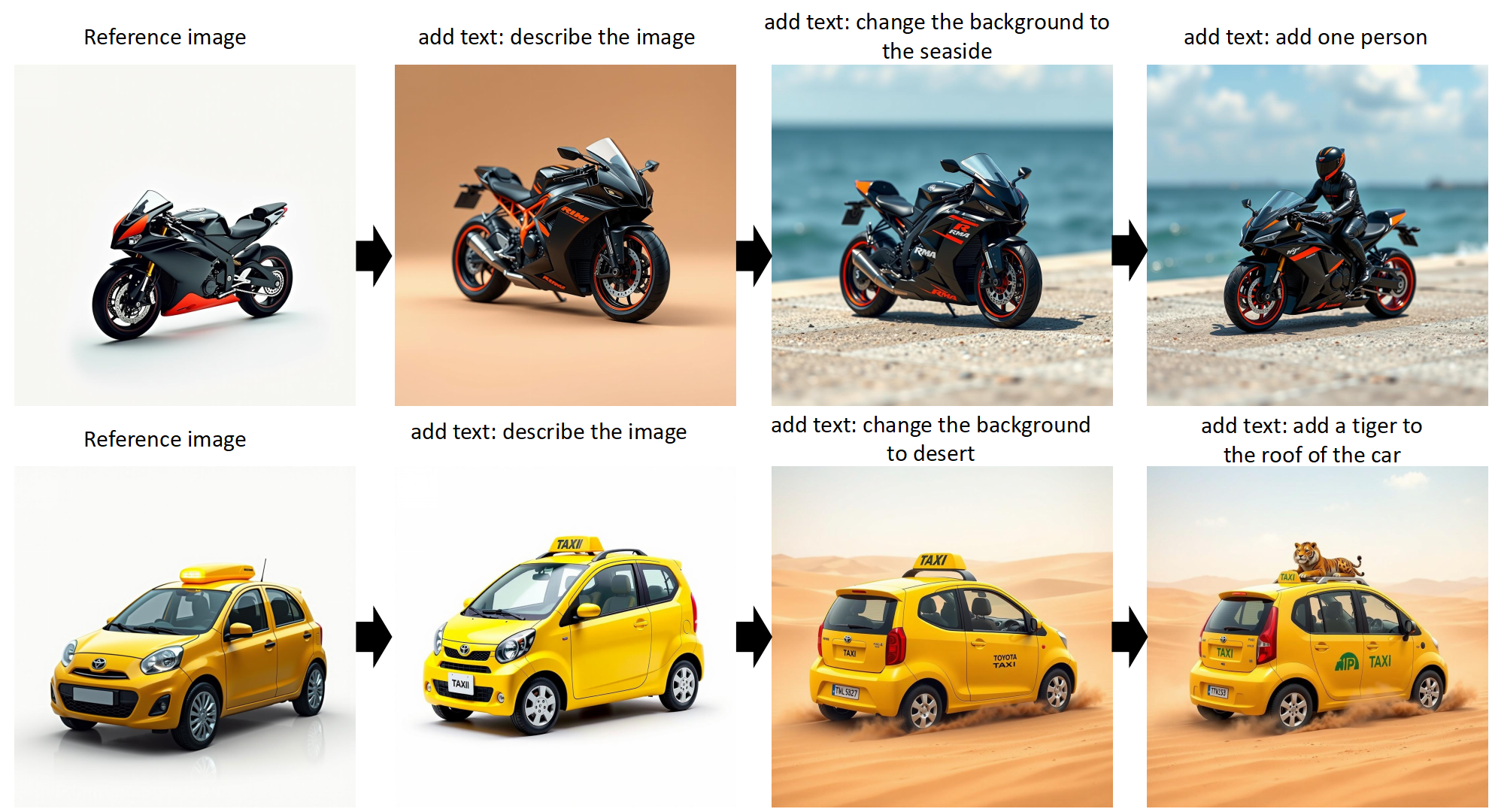}}
     \caption{X2I's multi-turn dialogue generation ability.}
  \label{fig:Multi-turn}
\end{figure}

MLLMs themselves possess certain reasoning abilities and multi-turn dialogue capabilities. X2I also inherits some of these related abilities.  \cref{fig:reasoning} illustrates the reasoning abilities based on text as well as image-based reasoning abilities. It is important to note here that the features fed into X2I originate from the characteristics of MLLMs' answers. \cref{fig:Multi-turn} demonstrates X2I's multi-turn dialogue capabilities, indicating that the generated images exhibit a certain degree of coherence and fidelity. Although these capabilities only represent a small portion of the abilities inherited from MLLMs, they provide more possibilities for improving the productivity of images.

\section{X2I Training with Different VLMs}
\label{sec:internVL}
\begin{figure*}[ht!]
	\centering
    \includegraphics[width=1\textwidth]{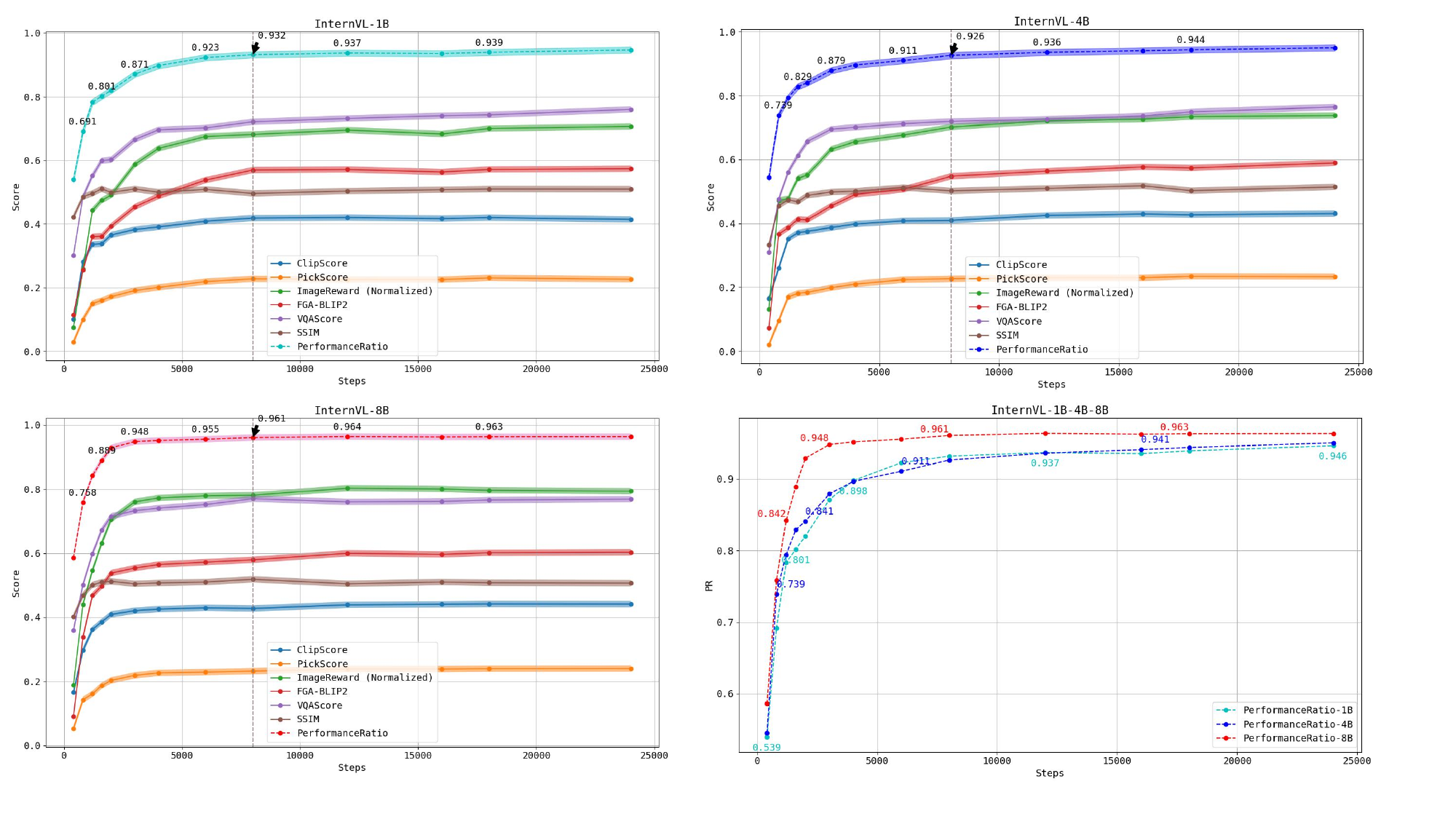}
     \caption{The Impact of varying capacities of VLMs on X2I training.}
  \label{fig:internVL}
\end{figure*}
\cref{fig:internVL} presents a comparison of experiments adapting X2I using three different model capacities based on InternVL. The top left shows results based on the InternVL-1B model, the top right is based on InternVL-4B, the bottom left is based on InternVL-8B, and the bottom right compares the performance of the three models on PR metrics. Overall, there are two conclusions to be drawn: firstly, as the capacity of the InternVL model increases, the peak value of the comprehensive X2I metrics also rises, which to some extent reflects the scaling-up capability of the model. The second conclusion is that as the model capacity increases, the fitting speed of X2I training also improves to some extent, likely benefiting from the design of AlignNet and the support of attention distillation.

\section{Consistency Analysis of Feature Distribution for Multiple Modes of MLLMs}
\label{sec:feature}
\begin{figure}[ht!]
	\centering
    \includegraphics[width=0.5\textwidth]{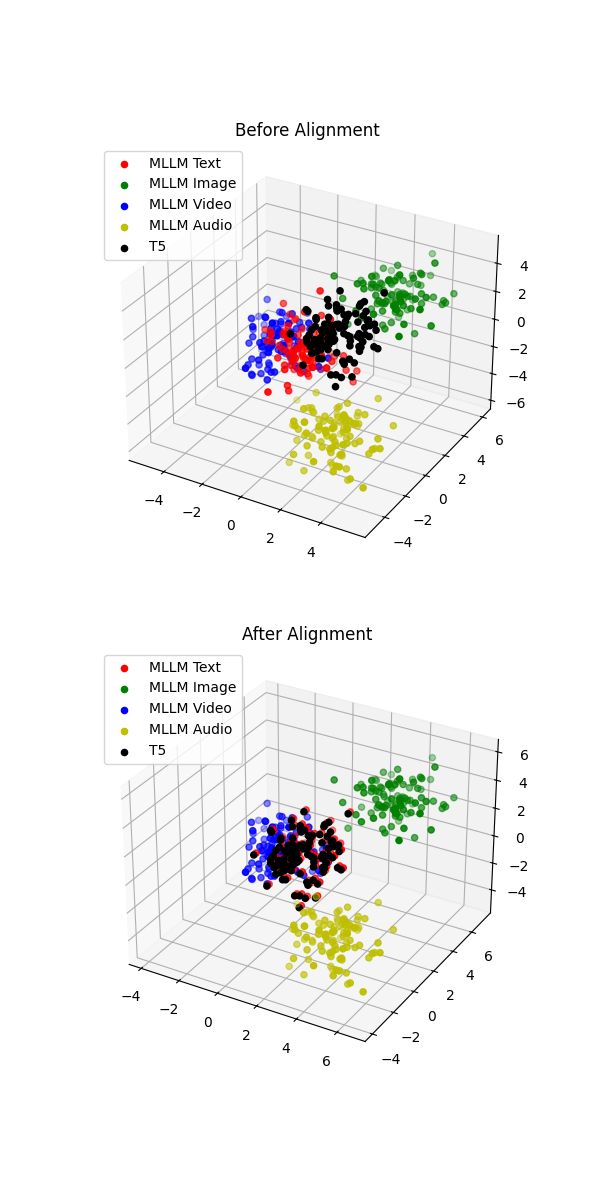}
     \caption{Schematic diagram of alignment process for different modalities.}
  \label{fig:feature}
\end{figure}

As shown in \cref{fig:feature}, in the unaligned state, MLLMs generate heterogeneous feature distributions for text, image, audio, and video across four modalities, embedding these distributions into a unified cross-modal shared semantic space. At this stage, there is a significant domain gap between the feature distribution of the text modality and the output of T5. By performing parameter space projection on the textual output layer of MLLMs through pure text training, the textual feature distribution is partially aligned with the T5 encoder. The aligned X2I architecture inherits the multimodal association capabilities of MLLMs via the shared semantic space.

\section{Acceleration of Training}
\label{sec:Acceleration}
Training-Inference Separation: Placing both the teacher model and the student model on the same GPU can lead to insufficient single-GPU memory, necessitating the use of ZeRO3\cite{rajbhandari2020zeromemoryoptimizationstraining}. However, ZeRO3 significantly reduces the inference speed of the teacher model and also decreases the training efficiency across multiple machines. By separating training and inference, the model that requires training and the model that performs only inference are placed on different GPUs. This setup results in very high inference efficiency for the teacher model without the need for multi-GPU communication. Meanwhile, the student model only requires data parallelism, with communication overhead being much lower than ZeRO3, greatly enhancing multi-machine training efficiency compared to ZeRO3.

Training-Inference Overlap: By overlapping the inference of the teacher model with the training of the student model, the inference time of the teacher model can be perfectly hidden. This approach significantly improves the efficiency of distillation training.

\section{Objective Performance of GenAI Bench and EvalMuse-40K}
\label{sec:GenAI}
\cref{tab:Universal2} displays the results from the GenAI-Bench and EvalMuse evaluation sets, with performance scores also exceeding 99\%. This further demonstrates that X2I's results in T2I generation are nearly indistinguishable from those of the teacher model Flux.1.

\begin{table*}[ht!]
\resizebox{1\linewidth}{!}{
\begin{tabular}{ccccccccccc}
\hline
Evaluation                    & Model  & ClipScore & PickScore & HPSv2  & ImageReward & FGA-BLIP2 & GPT-4o & VQAScore & BLIP-VQA & Performance              \\ \hline
\multirow{2}{*}{GenAI-Bench}  & Flux.1 & 0.4633    & 0.2907    & 0.2355 & 1.3105      & 3.6218    & 0.8195 & 0.7817   & 0.4970   & \multirow{2}{*}{99.13\%} \\
                              & X2I    & 0.4571    & 0.2873    & 0.2458 & 1.2536      & 3.5736    & 0.7925 & 0.7657   & 0.4957   &                          \\ \hline
\multirow{2}{*}{EvalMuse-40K} & Flux.1 & 0.4554    & 0.2575    & 0.2128 & 1.1721      & 3.4441    & 0.7980 & 0.7168   & 0.3627   & \multirow{2}{*}{99.09\%} \\
                              & X2I    & 0.4424    & 0.2482    & 0.2134 & 1.1315      & 3.4648    & 0.7718 & 0.7038   & 0.3559   &                          \\ \hline
\end{tabular}}
\caption{Objective Performance on the GenAI-Bench and EvalMuse Evaluation Dataset.}
\label{tab:Universal2}
\end{table*}

\section{Multilingual T2I Generation}
\label{sec:Multilingual}

\cref{tab:Multilingual} includes three multilingual evaluation datasets from Multilingual-General, as well as three additional translations in German, Portuguese, Spanish, and French. While the performance in English remains consistent with a one percentage point difference similar to before, the performance in other languages varies with some degree of decline. There are two main reasons for this difference:

The training corpus for QwenVL is primarily sourced from Chinese and English, with relatively fewer resources for other languages, making alignment with English datasets challenging.
Errors introduced by the translation software, as all eight evaluation metrics are tested on English evaluation datasets, necessitating a translation back to English for non-English languages. In comparing with existing open-source models in the industry, such as PEA-Diffusion and Sana, the multilingual T2I models in this study demonstrate significant advantages. PEA-Diffusion is based on the SDXL model, while our comparison involves the official open-source ``MultilingualFLUX.1". Sana is compared with the official open-source ``Sana\_1600M\_1024px\_MultiLing", with most of the Chinese T2I generation results failing, hence the lack of presentation of objective results for Chinese.

\begin{table*}[ht!]
\resizebox{1\linewidth}{!}{
\begin{tabular}{ccccccccccc}
\hline
Language                   & Model         & ClipScore & PickScore & HPSv2  & ImageReward & FGA-BLIP2 & GPT-4o & VQAScore & BLIP-VQA & Performance                             \\ \hline
                           & Flux.1        & 0.4517    & 0.2578    & 0.2433 & 1.3891      & 3.7789    & 0.9105 & 0.8741   & 0.4101   & -                                       \\ \cline{2-11} 
                           & PEA-Diffusion & 0.4325    & 0.2485    & 0.2401 & 1.1584      & 3.5987    & 0.8712 & 0.8501   & 0.3602   & 96.90\%                                 \\ \cline{2-11} 
\multirow{-3}{*}{English}  & X2I           & 0.4543    & 0.2577    & 0.2445 & 1.4043      & 3.7568    & 0.8905 & 0.8667   & 0.3905   & {\color[HTML]{24292F} \textbf{99.25\%}} \\ \hline
                           & PEA-Diffusion & 0.4252    & 0.2415    & 0.2412 & 1.0582      & 3.5542    & 0.8542 & 0.8469   & 0.3312   & 95.67\%                                 \\ \cline{2-11} 
\multirow{-2}{*}{Chinese}  & X2I           & 0.4481    & 0.2552    & 0.2429 & 1.2998      & 3.6458    & 0.8614 & 0.8601   & 0.3471   & \textbf{97.65\%}                        \\ \hline
                           & PEA-Diffusion & 0.4101    & 0.2202    & 0.2417 & 0.9854      & 3.2546    & 0.8498 & 0.8201   & 0.3042   & 93.33\%                                 \\
                           & Sana          & 0.4154    & 0.1929    & 0.2315 & 0.4168      & 2.9490    & 0.8045 & 0.7954   & 0.2855   & -                                       \\ \cline{2-11} 
\multirow{-3}{*}{German}   & X2I           & 0.4356    & 0.2368    & 0.2419 & 1.1736      & 3.5350    & 0.8715 & 0.8412   & 0.3178   & \textbf{96.03\%}                        \\ \hline
                           & PEA-Diffusion & 0.3952    & 0.2196    & 0.2399 & 1.0123      & 3.4550    & 0.7854 & 0.7952   & 0.3020   & 92.68\%                                 \\
                           & Sana          & 0.4253    & 0.1958    & 0.2217 & 0.5240      & 3.4910    & 0.7905 & 0.8014   & 0.3125   & -                                       \\ \cline{2-11} 
\multirow{-3}{*}{Portugal} & X2I           & 0.4344    & 0.2385    & 0.2427 & 1.1610      & 3.5892    & 0.8221 & 0.8254   & 0.3345   & \textbf{95.57\%}                        \\ \hline
                           & PEA-Diffusion & 0.4054    & 0.2285    & 0.2347 & 0.8975      & 3.2541    & 0.8014 & 0.8077   & 0.2998   & 92.20\%                                 \\
                           & Sana          & 0.4346    & 0.2080    & 0.2301 & 0.7202      & 3.1522    & 0.7899 & 0.8158   & 0.3101   & -                                       \\ \cline{2-11} 
\multirow{-3}{*}{Franch}   & X2I           & 0.4322    & 0.2353    & 0.2415 & 1.0956      & 3.4681    & 0.8317 & 0.8298   & 0.3210   & \textbf{94.91\%}                        \\ \hline
                           & PEA-Diffusion & 0.4051    & 0.2142    & 0.2298 & 1.0214      & 3.2121    & 0.8012 & 0.7946   & 0.3014   & 92.06\%                                 \\
                           & Sana          & 0.4199    & 0.1977    & 0.2145 & 0.5710      & 3.0526    & 0.7854 & 0.7714   & 0.2845   & -                                       \\ \cline{2-11} 
\multirow{-3}{*}{Spain}    & X2I           & 0.4320    & 0.2363    & 0.2427 & 1.1737      & 3.5398    & 0.8223 & 0.8399   & 0.3311   & \textbf{95.36\%}                        \\ \hline
                           & PEA-Diffusion & 0.4012    & 0.2154    & 0.2314 & 0.9214      & 3.2985    & 0.7988 & 0.7901   & 0.2952   & 91.84\%                                 \\
                           & Sana          & 0.3641    & 0.1479    & 0.2157 & -0.2139     & 2.5536    & 0.7012 & 0.7265   & 0.2545   & -                                       \\ \cline{2-11} 
\multirow{-3}{*}{Russia}   & X2I           & 0.4312    & 0.2354    & 0.2428 & 1.1475      & 3.5254    & 0.8237 & 0.8412   & 0.3236   & \textbf{94.94\%}                        \\ \hline
                           & PEA-Diffusion & 0.3925    & 0.2101    & 0.2351 & 0.9215      & 3.2541    & 0.7952 & 0.7892   & 0.2912   & 91.47\%                                 \\
                           & Sana          & 0.4124    & 0.1886    & 0.2201 & 0.4082      & 2.9347    & 0.7524 & 0.7412   & 0.2641   & -                                       \\ \cline{2-11} 
\multirow{-3}{*}{Itali}    & X2I           & 0.4286    & 0.2327    & 0.2419 & 1.0956      & 3.4769    & 0.8285 & 0.8214   & 0.3183   & \textbf{94.69\%}                        \\ \hline
\end{tabular}}
\caption{Multilingual T2I Performance on the Multilingual-General Evaluation Dataset.}
\label{tab:Multilingual}
\end{table*}

\section{IT2I and I2I Comparison of Objective Indicators}
\label{sec:IT2I}

We evaluate the ClipScore, PickScore, and ImageReward. The input prompts consist of the reference image prompt combined with editing prompts. For an objective test, we focus on common editing tasks such as addition, specifically adding five common animals (cat, dog, lion, tiger, elephant) to the original image. This evaluation primarily assesses the image instruction editing ability.
For I2I metrics, we also provide SSIM scores. The corresponding prompt is the prompt of the reference image itself. The reference images are generated by the teacher model corresponding to 200 English corpus from the Multilingual-General dataset. This evaluation mainly examines the model's capability to generate variations of images. The baseline model for comparison is Qwen2VL-Flux.

\textbf{IT2I and I2I}. \cref{tab:IT2I} presents metrics for IT2I and I2I, comparing the X2I model with Qwen2VL-Flux. In all results, the metrics of X2I are significantly higher than those of Qwen2VL-Flux, indicating that X2I have a strong semantic understanding of images and performs well in common image generation tasks involving instruction editing.

\begin{table*}[ht!]
\centering 
\resizebox{0.8\linewidth}{!}{
\begin{tabular}{cccccccc}
\hline
\multirow{3}{*}{Model} & \multicolumn{4}{c}{I2I}       & \multicolumn{3}{c}{TI2I}     \\ \cline{2-8} 
                       & \multirow{2}{*}{ClipScore} & \multirow{2}{*}{PickScore} & \multirow{2}{*}{ImageReward} & \multirow{2}{*}{SSIM} & \multirow{2}{*}{ClipScore} & \multirow{2}{*}{PickScore} & \multirow{2}{*}{ImageReward} \\
                       &                            &                            &                              &                       &                            &                            &                              \\ \hline
Qwen2VL-Flux              & 0.3919                     & 0.1845                     & 0.6824                       & 0.4045                & 0.3772                     & 0.1429                     & -0.3914                      \\ \hline
X2I                    & \textbf{0.4347}            & \textbf{0.2371}            & \textbf{1.2446}              & \textbf{0.5208}       & \textbf{0.4318}            & \textbf{0.182}             & \textbf{0.732}               \\ \hline
\end{tabular}}
\caption{Performance of X2I in IT2I and I2I Generation.}
\label{tab:IT2I}
\end{table*}

\section{Plug-and-Play}
\label{sec:Downstream}

Benefiting from the alignment strategy of attention distillation, the MLLM encoder acquired by X2I can be directly transferred to Flux.1 variant models and downstream ecosystems without requiring retraining.
\subsection{Transfer Capability on Flux.1 Variant Models} 
By replacing the text encoders of the sampling acceleration model Flux.1-Turbo, aesthetic fine-tuning model Flux.1-Shuttle3.1, and compression model FLEX with X2I's MLLM as the encoder, we derived corresponding variants X2I-Turbo, X2I-Shuttle, and X2I-FLEX. These variants are systematically compared with the original models in text-to-image generation performance. As demonstrated in \cref{fig25}, \cref{fig26}, and \cref{fig27}, our alignment strategy achieves robust cross-architecture transferability.

\subsection{Downstream Task Transferability} 
The aligned MLLM encoder can be directly transferred to downstream ecosystems including LoRA, inpainting, outpainting, ControlNet, and IP-Adapter without architectural modifications. As demonstrated in \cref{fig28}, \cref{fig29}, and \cref{fig30}, X2I achieve comparable or superior performance on children's simple sketch LoRA, outpainting, and inpainting tasks.

\textbf{ControlNet}. 
For ControlNet-related tasks, X2I's utilization of the MLLM as an encoder enables original image input support, thereby capturing finer-grained visual details. As shown in \cref{fig31}, X2I achieves enhanced performance under canny, depth, and hed reference conditions.  We also compared the performance of ControlNet downstream models based on Flux.1 directly applied to the X2I framework as \cref{tab:ControlNet} shows. X2I w/i indicates that the reference image is also input to the X2I framework during inference. The results show that including the image can further enhance the performance on relevant tasks based on three metrics.
\begin{table}[ht!]
\centering
\resizebox{0.8\linewidth}{!}{
\begin{tabular}{llll}
\hline
ControlNet & \begin{tabular}[c]{@{}l@{}}Canny Edge\\ (F1 Score↑)\end{tabular} & \begin{tabular}[c]{@{}l@{}}Hed Edge\\ (SSIM↑)\end{tabular} & \begin{tabular}[c]{@{}l@{}}Depth Map\\ (RMSE↓)\end{tabular} \\ \hline
Original   & 0.2903 & 0.5025                                                     & 53.79                                                       \\ \hline
X2I        & 0.3144 & 0.5340                                                     & 49.09                                                       \\ \hline
X2I w/i    & \textbf{0.3304}                                                  & \textbf{0.5377}                                            & \textbf{43.03}                                              \\ \hline
\end{tabular}}
\caption{Performance of X2I Framework in Transfer to ControlNet Downstream Tasks.}
\label{tab:ControlNet}
\end{table}

\textbf{IP-Adapter}. 
In IP-Adapter tasks in \cref{fig32}, X2I exhibits improved fidelity preservation owing to its original image encoding capability. Similarly, we compared the performance of the IP-Adapter downstream model based on Flux.1 directly applied to the X2I framework as \cref{tab:IPAdapter} shows. The evaluation dataset used is DreamBench. Without including image inputs, X2I shows a certain advantage in DINO and CLIP-I metrics, but slightly decreases. X2I w/i, with image inputs, exhibits a significant advantage in DINO and CLIP-I, but sacrifices semantic adherence capabilities as seen from CLIP-T. Similar conclusions can be drawn from the comparison table.
\begin{table}[ht!]
\centering
\resizebox{0.8\linewidth}{!}{
\begin{tabular}{llll}
\hline
IP-Adapter & DINO           & CLIP-I         & CLIP-T         \\ \hline
Original  & 0.768          & 0.803          & \textbf{0.322} \\ \hline
X2I       & 0.779          & 0.811          & 0.320          \\ \hline
X2I w/i   & \textbf{0.863} & \textbf{0.865} & 0.308          \\ \hline
\end{tabular}}
\caption{Performance of X2I Framework in Transfer to IP-Adapter Downstream Tasks.}
\label{tab:IPAdapter}
\end{table}

\section{Template for X2I's input}
\label{sec:Template}
X2I have designed an input template for the student model during the alignment training process as follows:  \{``text prompt":``",``editing prompt":``",``image prompt":``"，,``video prompt":``"，,``audio prompt":``"\},The ``text prompt" represents the original prompt for generating images, the ``editing prompt" is the prompt for editing instructions for the input image, video, or audio, and the ``image prompt," ``video prompt," and ``audio prompt" respectively signify whether there is information from an input image, video, or audio. The motivation behind designing such a general template is to provide a paradigm input for subsequent stage training, LoRA training, and more controllable post-training.

\section{LoRA for IT2I}
\label{sec:LoRA}

Currently, the prevalent approach in the industry for generating visual images involves training models based on textual conditions as input. There is a lack of LoRA training for scenarios where the user's input involves highly personalized reference images for image-to-image and image-to-text generation. For instance, taking ``style information from the reference image alone" as an image edit prompt, the training of the general X2I framework primarily captures a global semantic understanding of the image, making it challenging to individually parse and handle finer-grained layered features of the image. To address this issue, we can train a conventional LoRA module within the X2I framework to solve these kinds of problems.

This paper trained the LoRA module for IT2I. We collected 100 abstract artistic sketches and created 100 data pairs with the input template \{``text prompt": ``", ``editing prompt": ``Style Information Only from Reference Image", ``image prompt": ``yes", ``video prompt": ``no", ``audio prompt": ``no"\}. Here, the text prompt is set as the corresponding textual description for each of the 100 images collected, and the ``image prompt" is randomly selected from the 100 data pairs.

\begin{figure*}[ht!]
	\centering
    \includegraphics[width=1\textwidth]{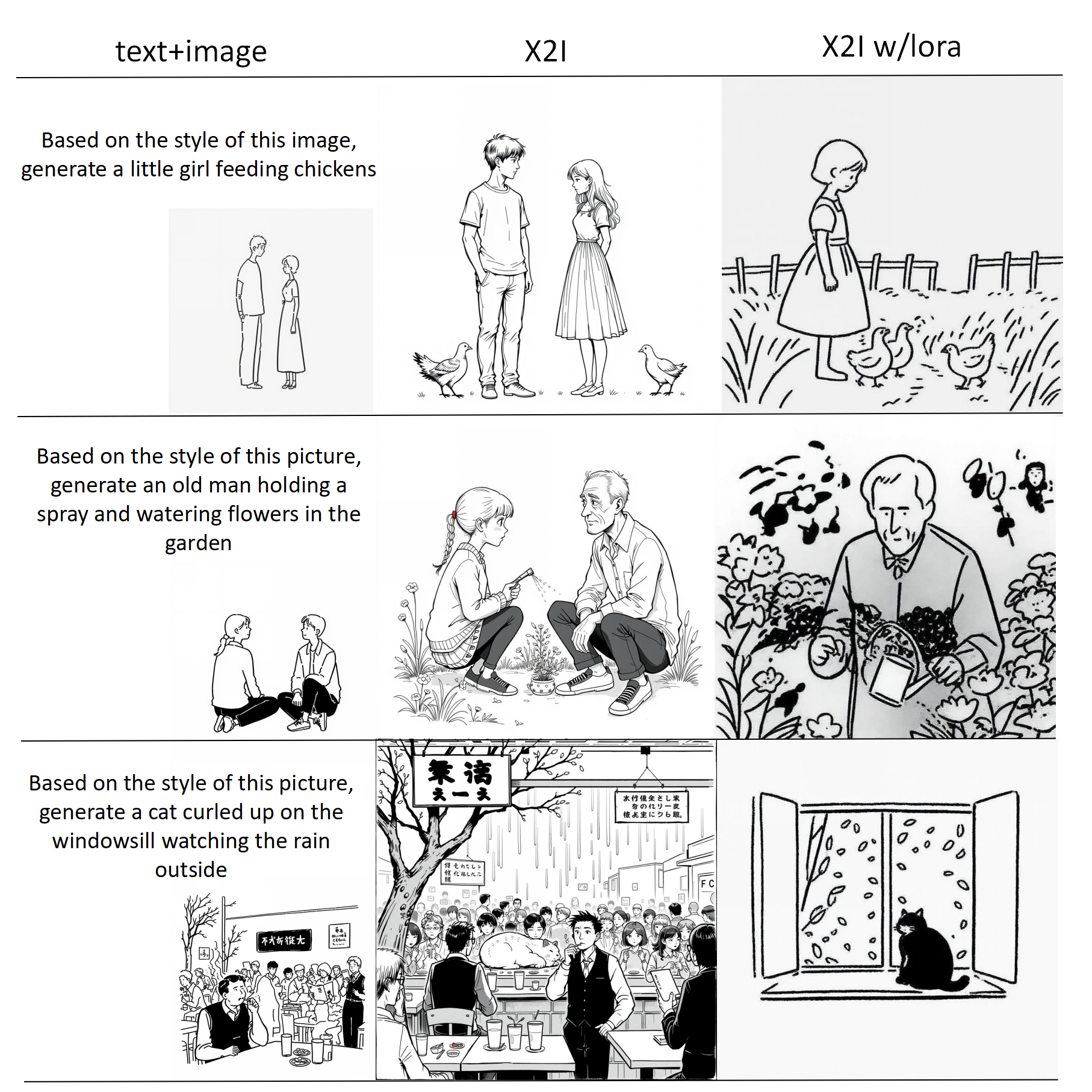}
     \caption{The effects before and after LoRA training. The conclusion indicates that LoRA training can encourage the X2I framework to selectively learn and extract different semantic information from input images.}
  \label{fig:lora}
\end{figure*}

\cref{fig:lora} shows the comparison of results before and after LoRA training. The conclusion indicates that LoRA training can prompt the X2I framework to selectively learn and extract different semantic information from input images.

\section{X2I enhanced Training Results}
\label{sec:style}

\begin{figure}[ht!]
	\centering
    \includegraphics[width=0.5\textwidth]{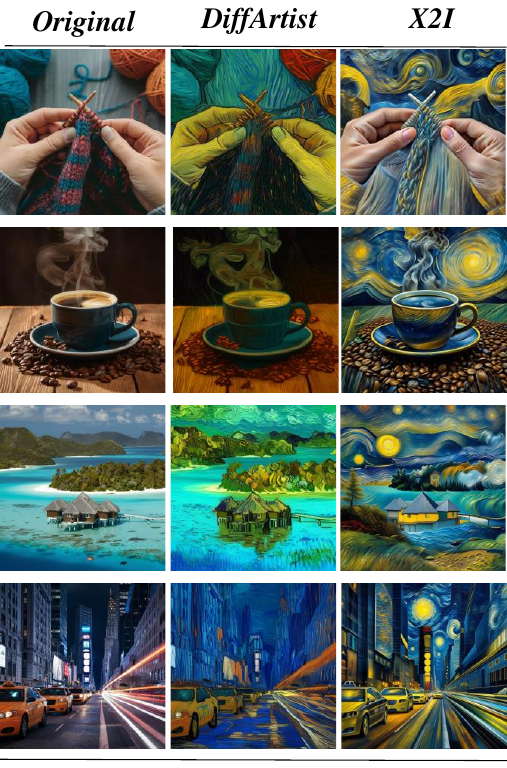}
     \caption{Comparison between DifffArtist and X2I Van Gogh style transfer capability.}
  \label{fig11}
\end{figure}

\begin{figure}[ht!]
	\centering
    \includegraphics[width=0.5\textwidth]{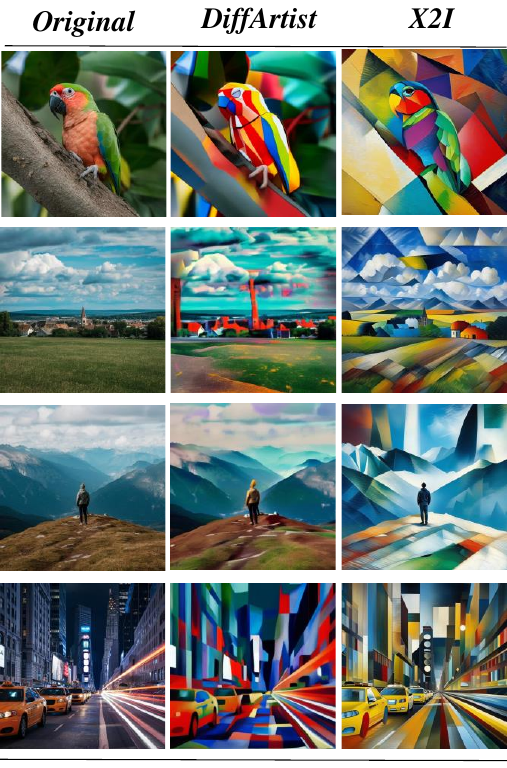}
     \caption{Comparison between DifffArtist and X2I cubism style transfer capability.}
  \label{fig12}
\end{figure}

We further present a comparison of Van Gogh style \cref{fig11} and Cubism style \cref{fig12}. Overall, the semantic consistency ability of X2I shows some improvement compared to DifffArtist, and the fidelity to the reference image significantly improves compared to X2I trained without enhancement.


\begin{figure*}[htbp]
    \centering
    \includegraphics[width=1\linewidth]{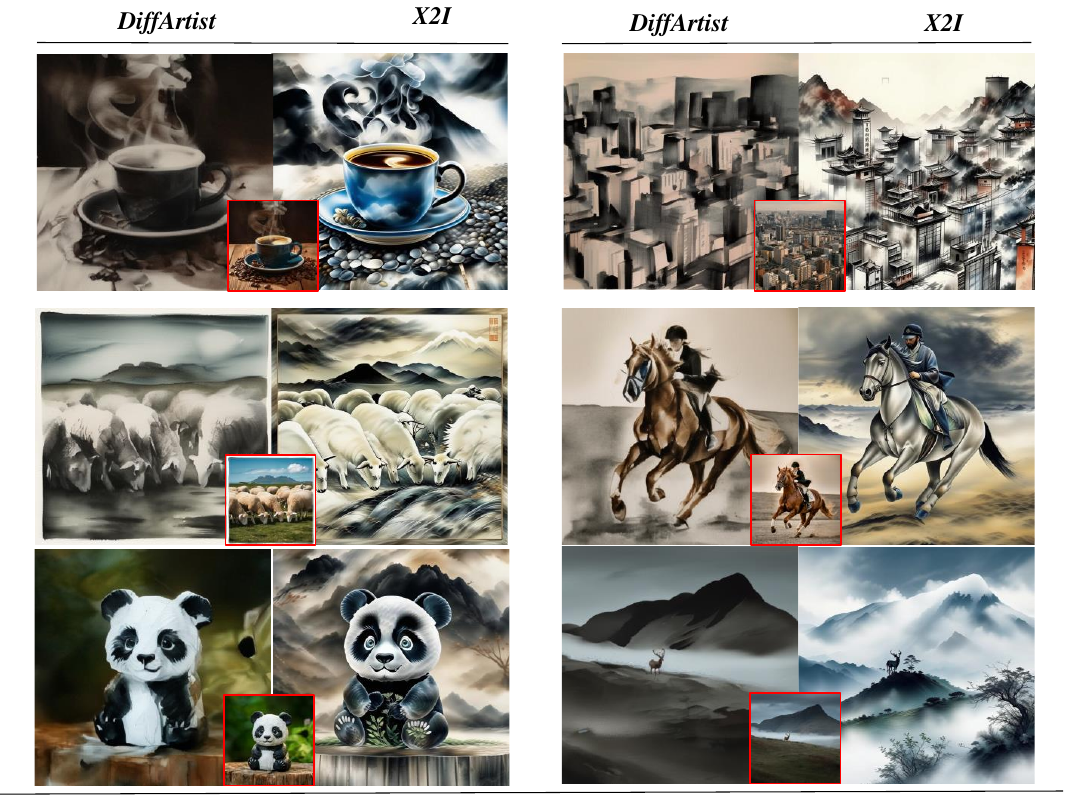}
    \caption{Comparison between DifffArtist and X2I ink painting style transfer capability.}
    \label{fig13}
\end{figure*}

\section{X2I Generation} 
\label{sec:zero}


\subsection{Text-to-Image} 
\begin{figure*}[htbp]
    \centering
    \includegraphics[scale=0.78]{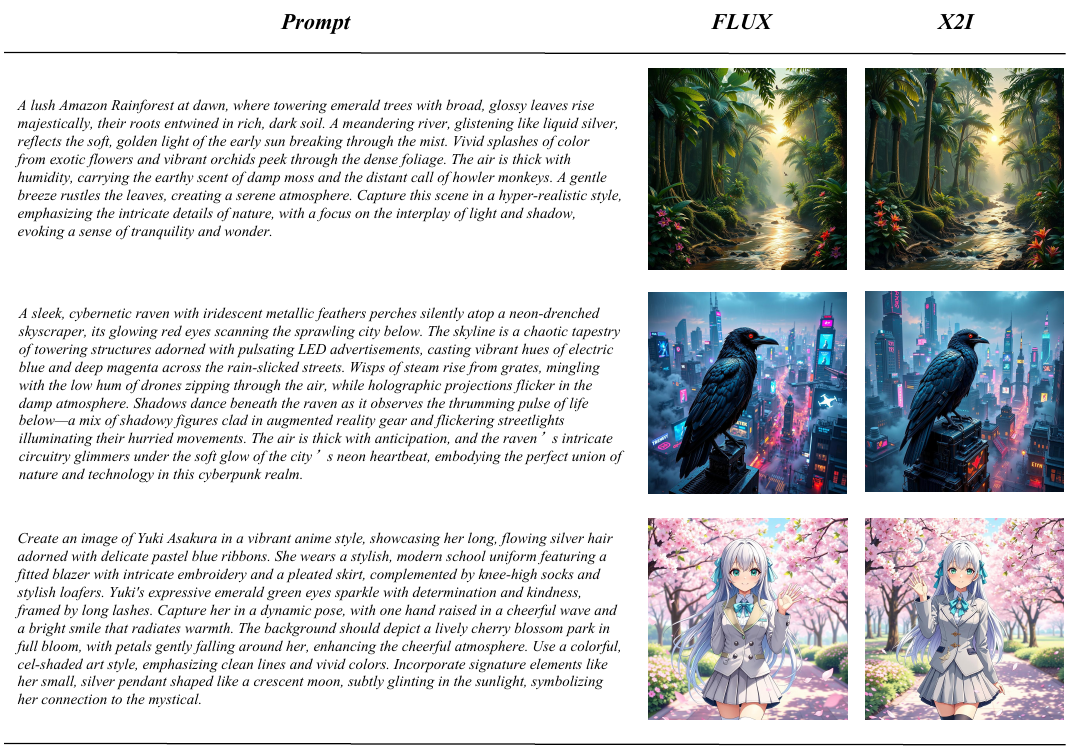}
    \caption{Comparison of the image generation results using the same prompt between the original FLUX.1 and our X2I.}
    \label{figure1}
\end{figure*}

\begin{figure*}[htbp]
    \centering
    \includegraphics[scale=0.78]{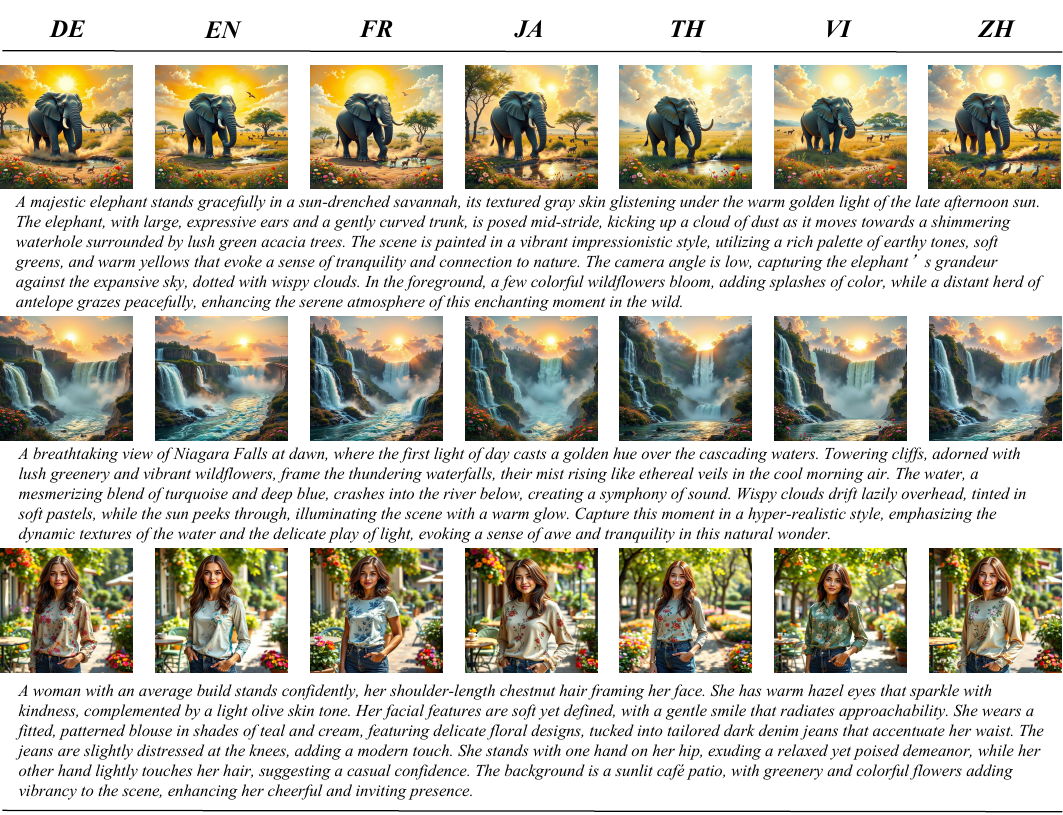}
    \caption{Comparison of text-to-image generation results using different languages.}
    \label{figure2}
\end{figure*}

\cref{figure1} compares the effects of Flux.1 and X2I in generating images for complex prompts , and also \cref{figure2} demonstrates X21's ability to generate images in multiple languages, including German, English, French, Japanese, Thai, Vietnamese, and Chinese.
\subsection{Image-to-Image} 
\begin{figure*}[htbp]
    \centering
    \begin{minipage}{0.48\linewidth} 
        \centering
        \includegraphics[scale=0.8]{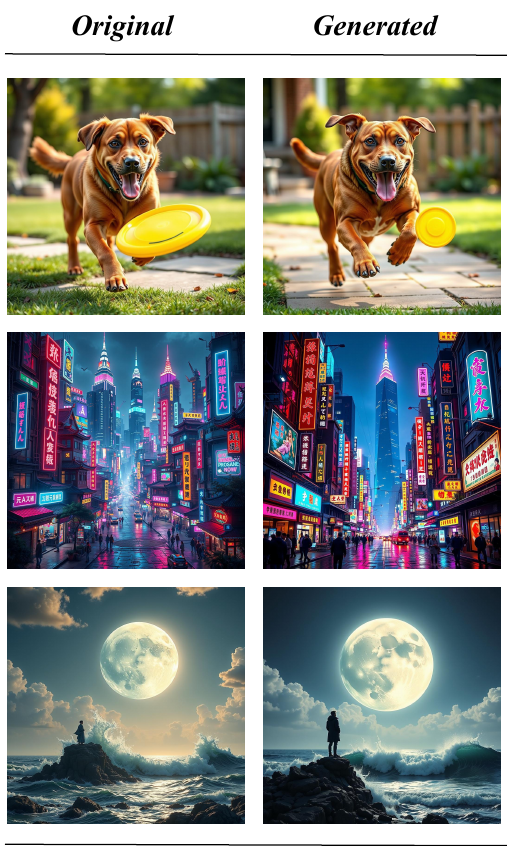}
        \caption{Image-to-image generation.}
        \label{fig3}
    \end{minipage}
    \hfill 
    \begin{minipage}{0.48\linewidth} 
        \centering
        \includegraphics[scale=0.8]{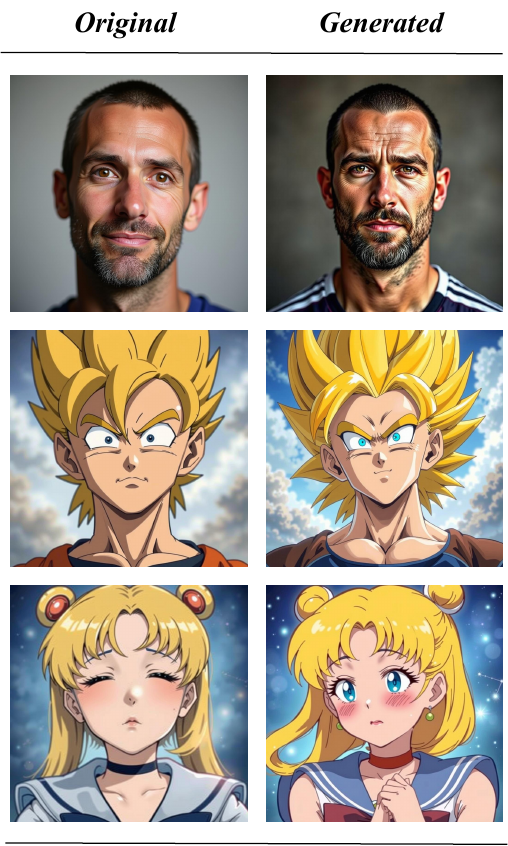}
        \caption{Celebrity or IP image-to-image generation.}
        \label{fig4}
    \end{minipage}
\end{figure*}

\begin{figure*}[htbp]
    \centering
    \begin{minipage}{0.48\linewidth} 
        \centering
        \includegraphics[scale=0.8]{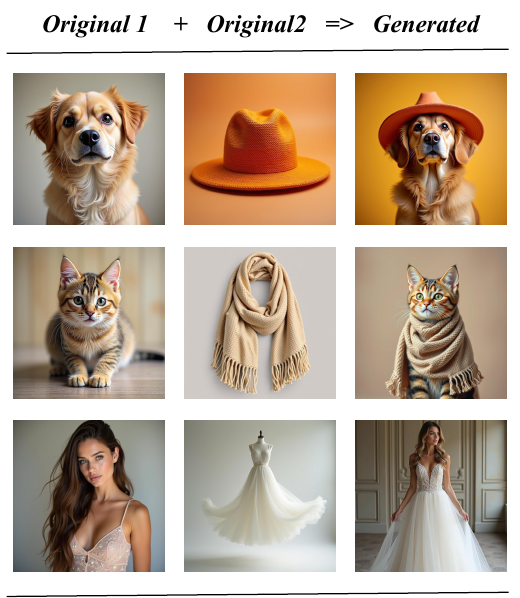}
        \caption{Clothing and accessories multi-image generation.}
        \label{fig5}
    \end{minipage}
    \hfill 
    \begin{minipage}{0.48\linewidth} 
        \centering
        \includegraphics[scale=0.8]{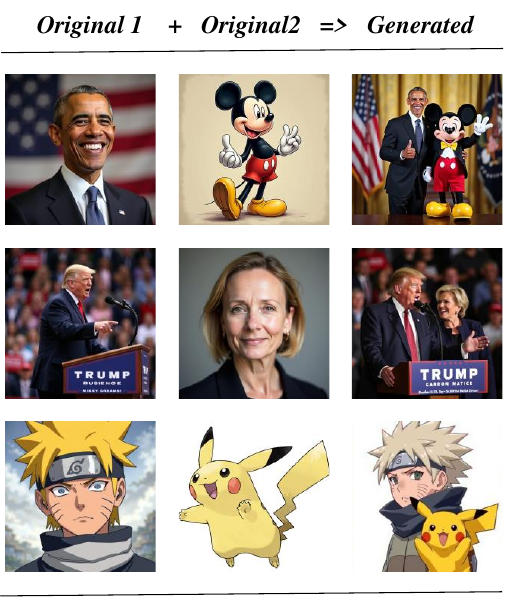}
        \caption{Group photo multi-image generation.}
        \label{fig6}
    \end{minipage}
\end{figure*}

\begin{figure*}[htbp]
    \centering
    \includegraphics[scale=0.76]{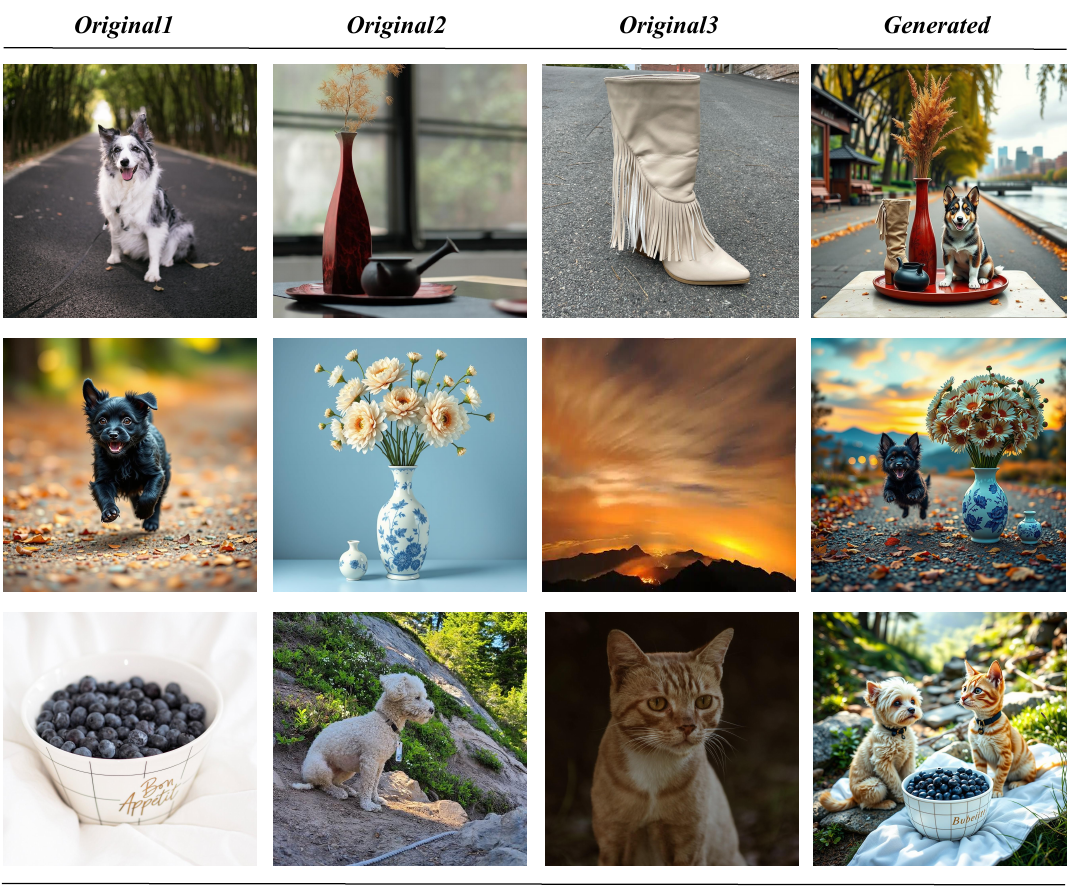}
    \caption{Blend three images to generate a new image.}
    \label{fig7}
\end{figure*}

\begin{figure*}[htbp]
    \centering
    \includegraphics[scale=0.76]{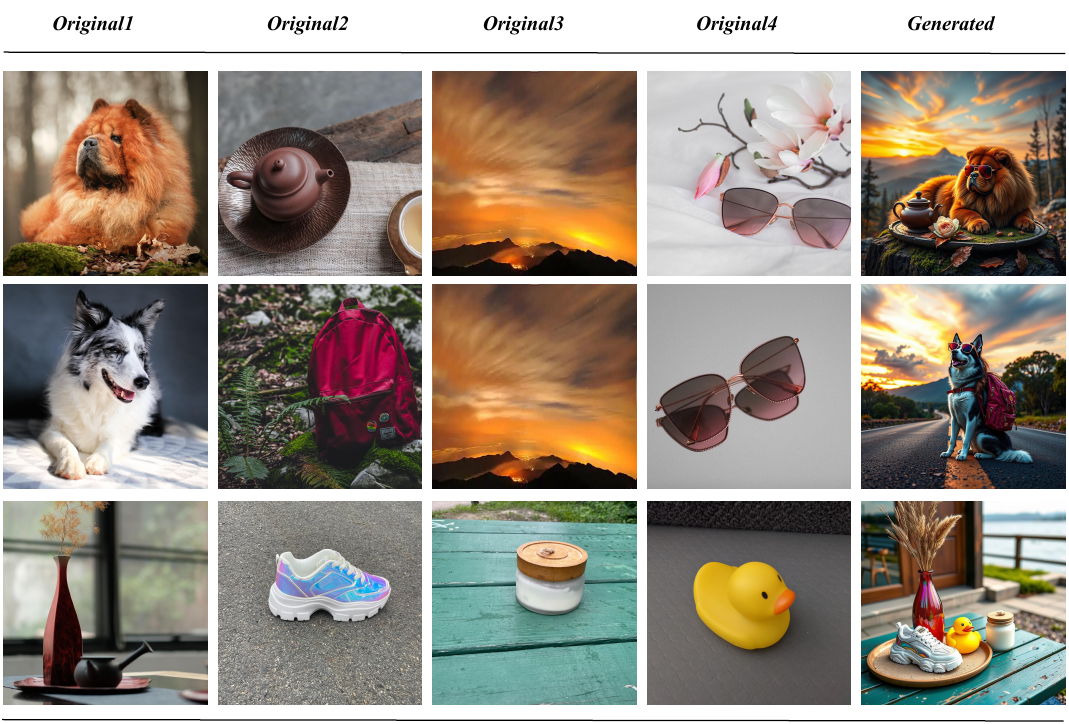}
    \caption{Blend four images to generate a new image.}
    \label{fig8}
\end{figure*}

MLLM empowers X2I with the capability to understand both single and multiple images, enabling it to perform reference-guided image generation in \cref{fig3}, celebrity and IP generation in \cref{fig4}, and multi-image composition tasks in \cref{fig5}, \cref{fig6}, \cref{fig7} and \cref{fig8}.
\subsection{Text+Image-to-Image} 
\begin{figure*}[htbp]
    \centering
    \includegraphics[scale=0.8]{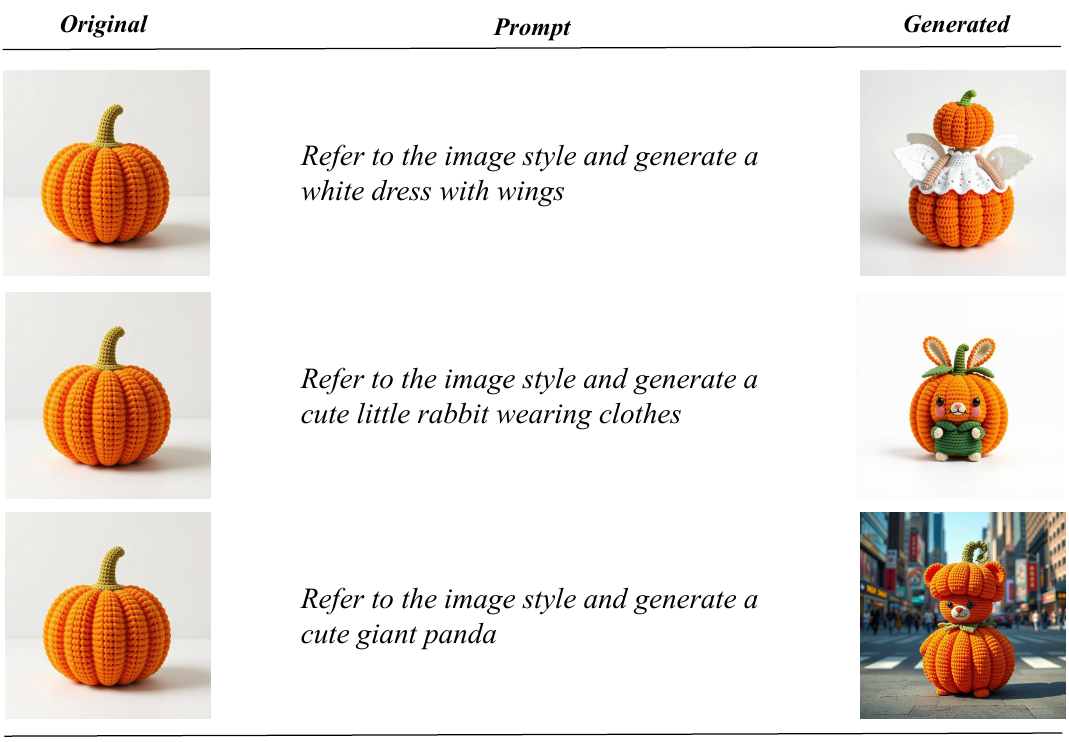}
    \caption{Image style transfer, refer to the image style and generate an image based on the prompt.}
    \label{fig9}
\end{figure*}

\begin{figure*}[htbp]
    \centering
    \includegraphics[scale=0.8]{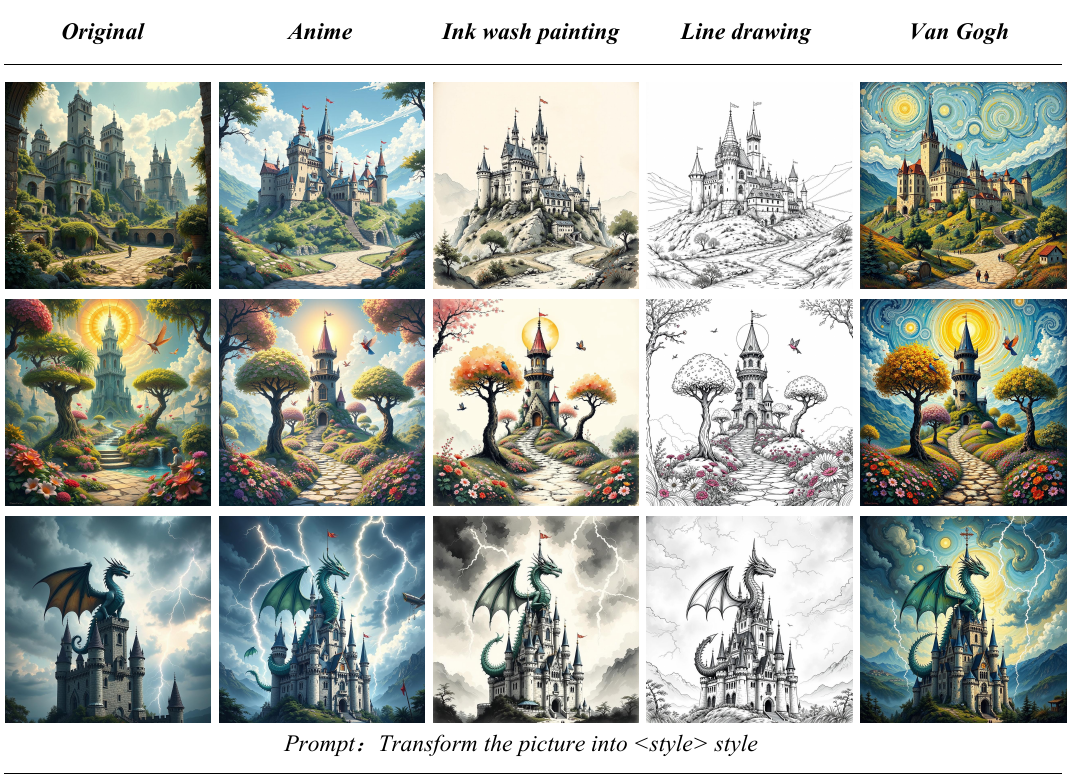}
    \caption{Style Editing, change the style of the specified image.}
    \label{fig10}
\end{figure*}

\begin{figure*}[htbp]
    \centering
    \includegraphics[scale=0.76]{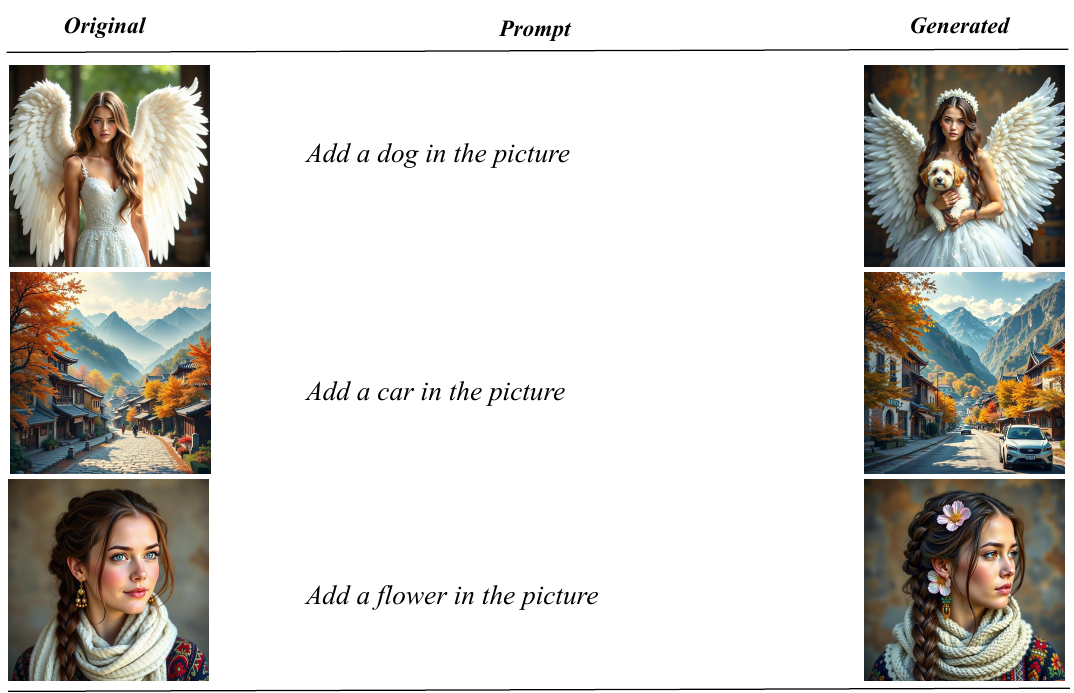}
    \caption{Image editing, add a new object to the image.}
    \label{fig14}
\end{figure*}

\begin{figure*}[htbp]
    \centering
    \includegraphics[scale=0.76]{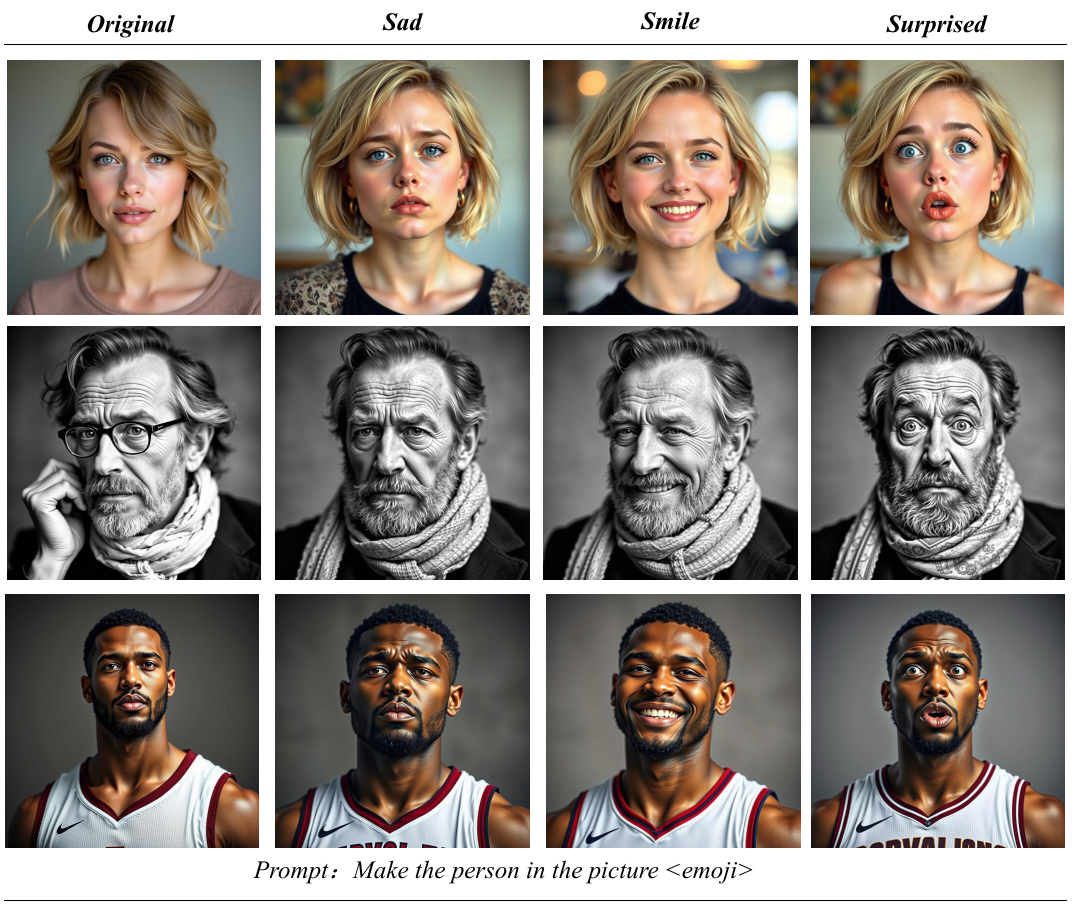}
    \caption{Expression editing, change the expression of a person in the image.}
    \label{fig15}
\end{figure*}

\begin{figure*}[htbp]
    \centering
    \includegraphics[scale=0.8]{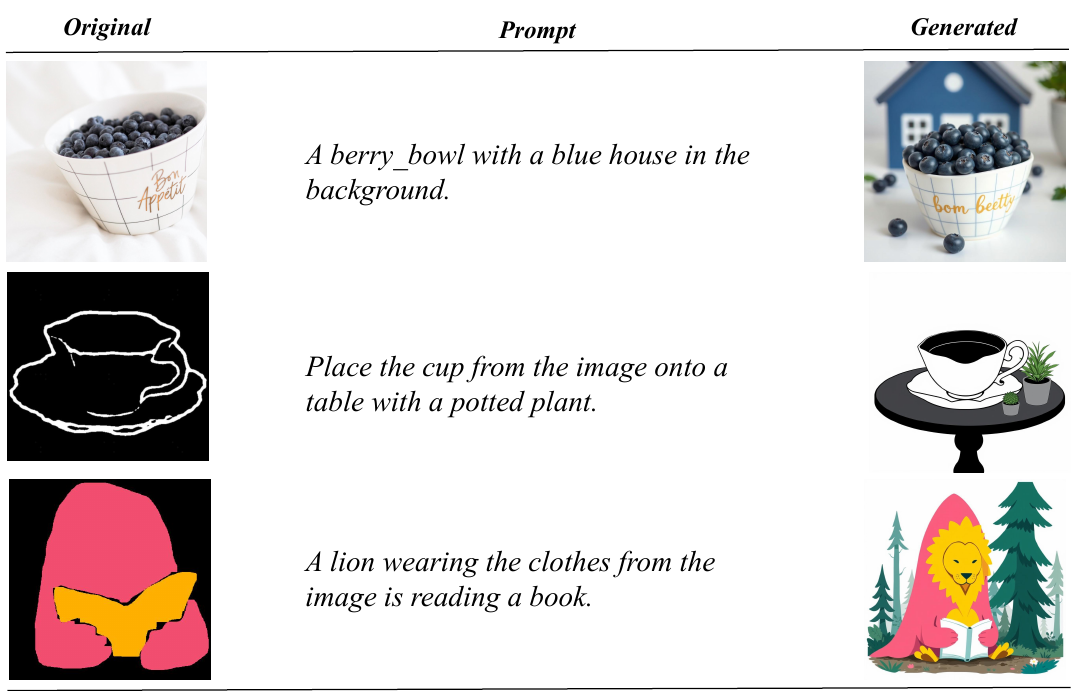}
    \caption{Single image and prompt fusion for image generation.}
    \label{fig16}
\end{figure*}

\begin{figure*}[htbp]
    \centering
    \includegraphics[scale=0.8]{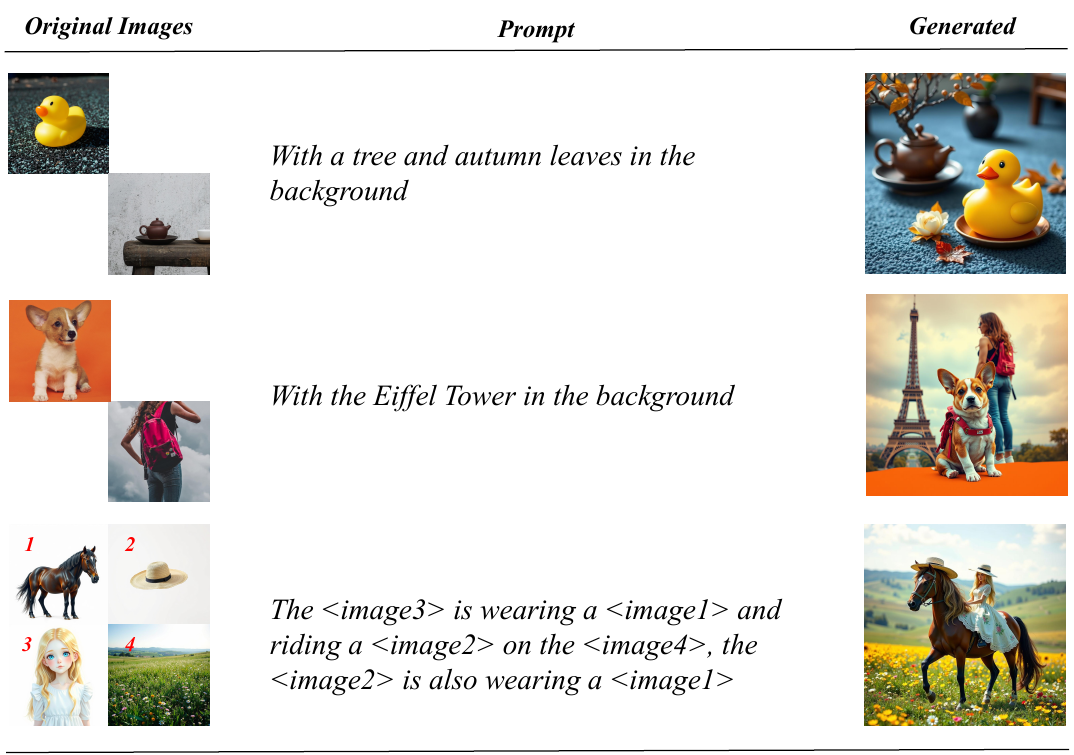}
    \caption{Multiple image and prompt fusion for image generation, in the third case, in order to express the complex relationship between multiple entities, we generate images by embedding image tokens between text.}
    \label{fig17}
\end{figure*}

\begin{figure*}[htbp]
    \centering
    \includegraphics[scale=1]{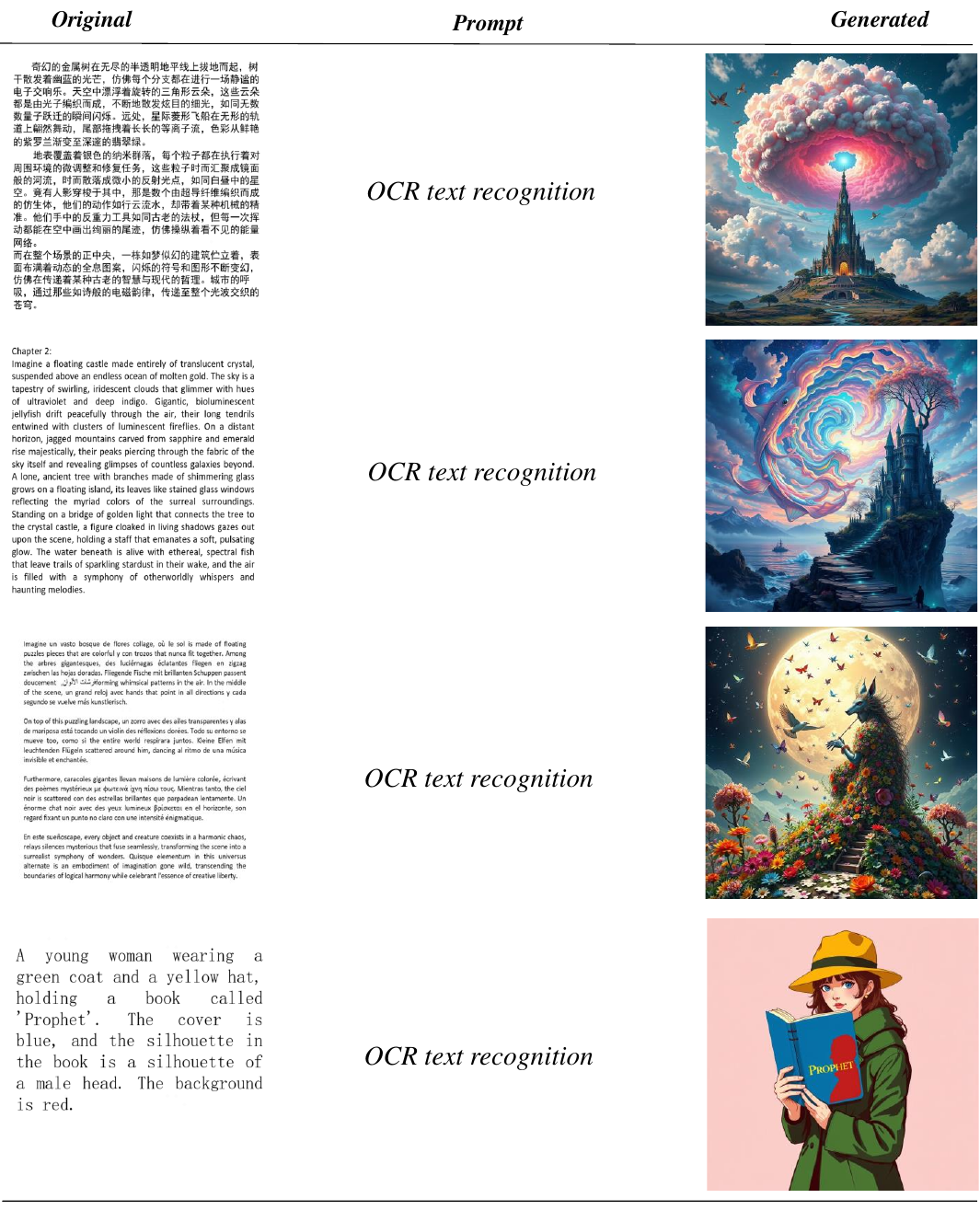}
    \caption{Image generation from screenshot inputs with OCR recognition.}
    \label{fig18}
\end{figure*}

X2I demonstrates capabilities including user-prompt-driven style transfer, expression editing as shown in \cref{fig9}, \cref{fig10} and \cref{fig15}, along with single image or multi-image editing and fusion tasks illustrated in \cref{fig14}, \cref{fig16}, and \cref{fig17}. Furthermore, leveraging MLLM's robust OCR capacity, the system generates images through direct interpretation of visual content in input images while supporting multilingual visual generation, evidenced by the results in \cref{fig18}.
\subsection{Video-to-Image} 
\begin{figure*}[htbp]
    \centering
    \includegraphics[scale=1]{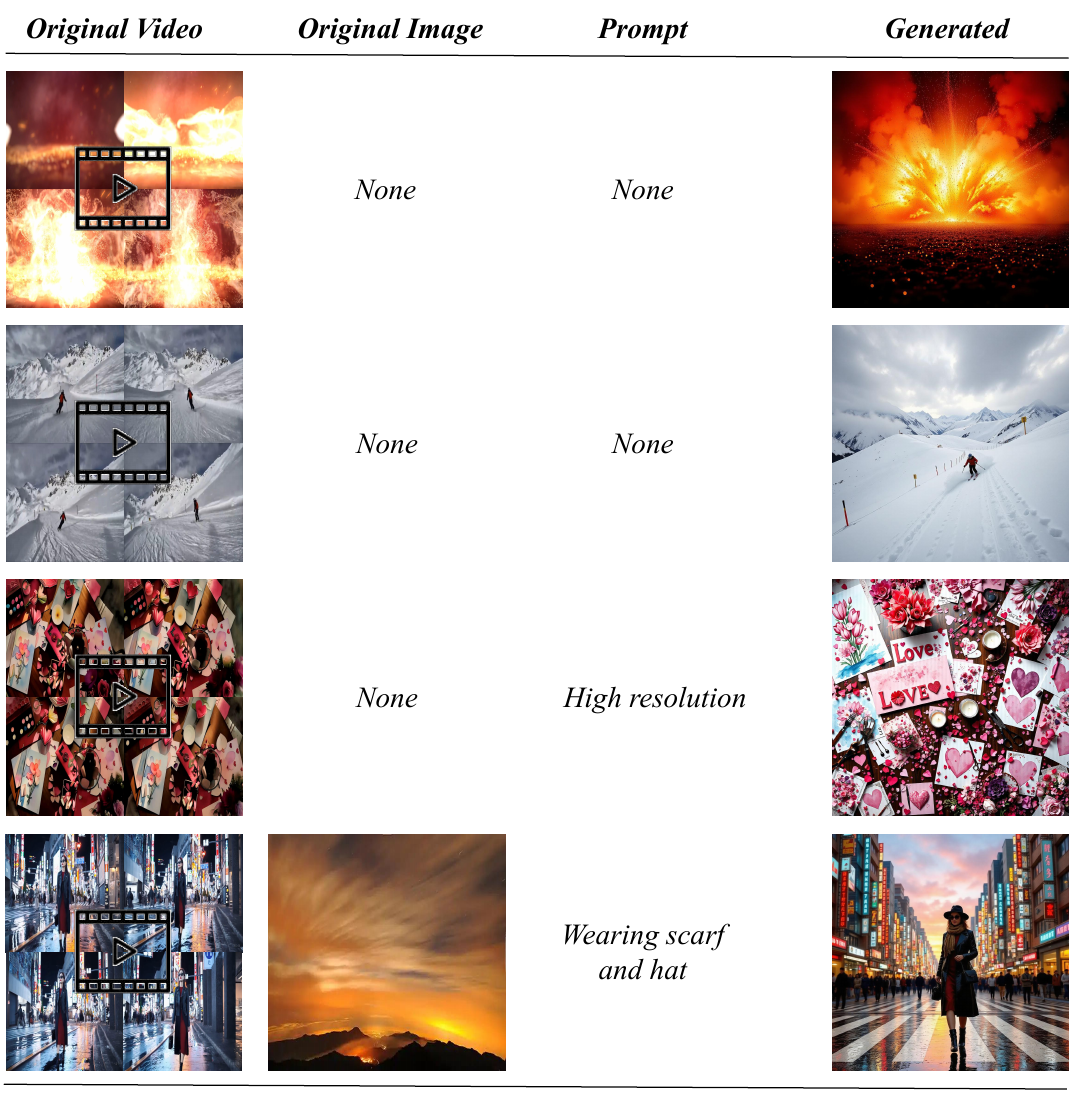}
    \caption{Video-to-image generation, generate images based on the content of the video.}
    \label{fig19}
\end{figure*}

MLLM possesses video comprehension capabilities that enable X2I to directly generate images based on the semantic content of input video sequences, as shown in \cref{fig19}. X2I can still generate high-resolution images through the prompt of the low-resolution video.
\subsection{Audio-to-Image} 
\begin{figure*}[htbp]
    \centering
    \includegraphics[scale=0.8]{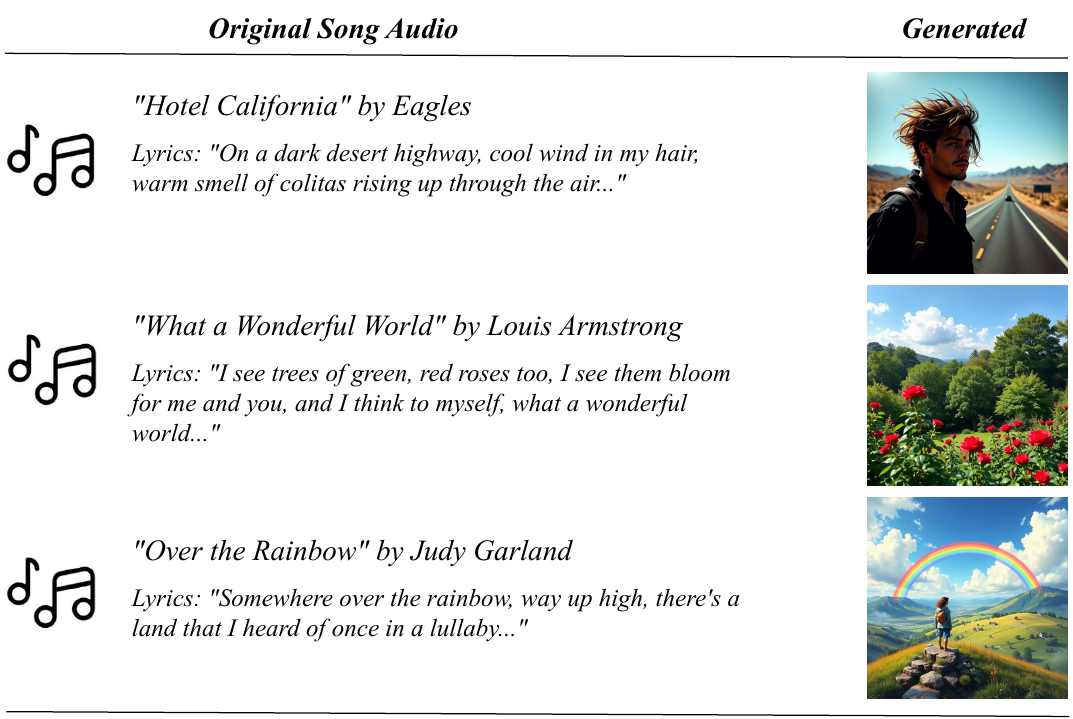}
    \caption{Music-to-image generation, generate images from music with lyrics.}
    \label{fig20}
\end{figure*}

\begin{figure*}[htbp]
    \centering
    \includegraphics[scale=0.8]{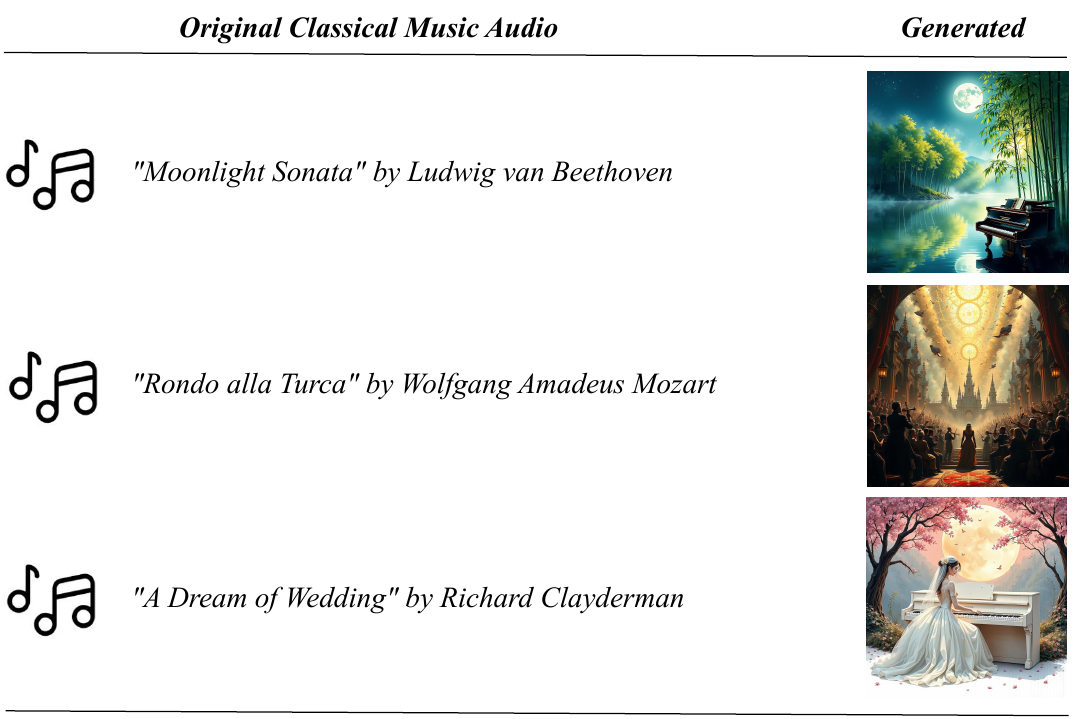}
    \caption{Instrumental-music-to-image generation, generate images from instrumental music audio.}
    \label{fig21}
\end{figure*}

\begin{figure*}[htbp]
    \centering
    \includegraphics[scale=0.75]{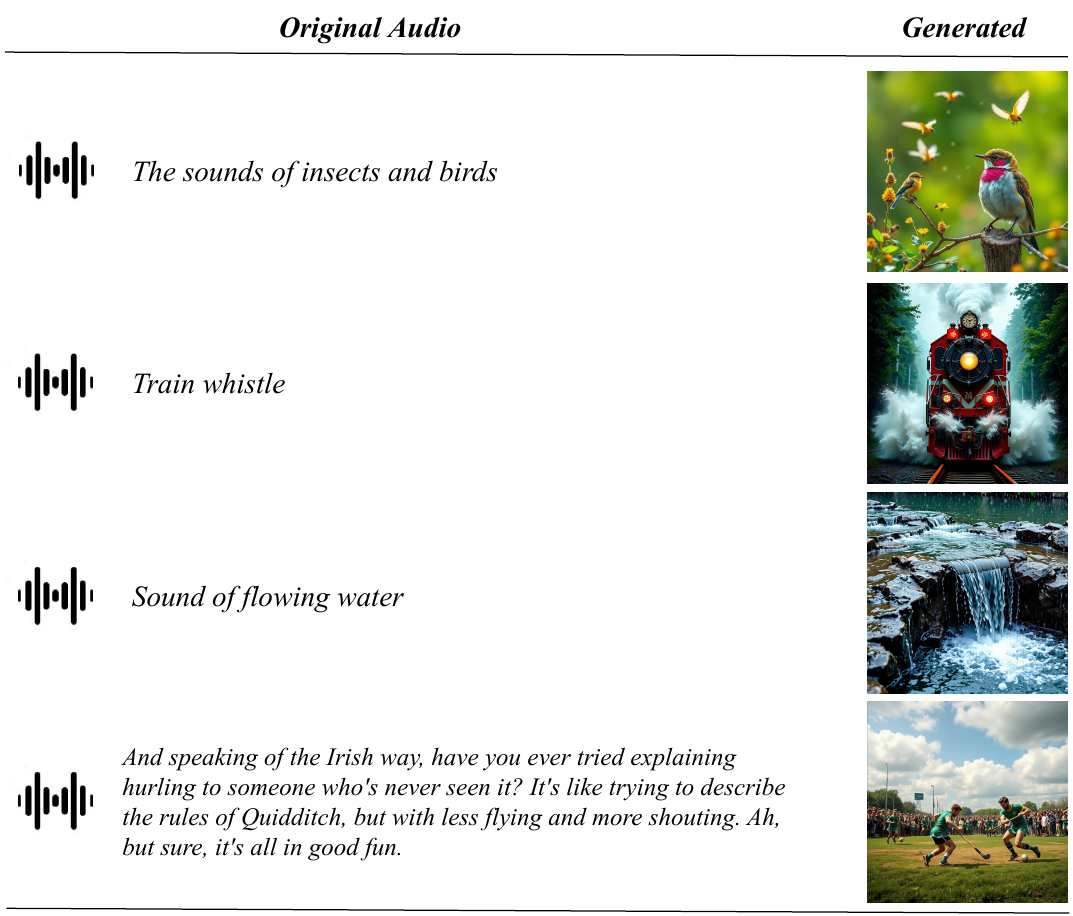}
    \caption{Audio-to-image generation, generate images from natural sounds, vehicle sounds, and complex human sounds.}
    \label{fig22}
\end{figure*}

Leveraging the audio comprehension capabilities of MLLMs such as MiniCPM-o, after alignment, X2I can directly generate images based on music with lyrics, instrumental music, and natural sounds as shown in \cref{fig20}, \cref{fig21} and \cref{fig22}. All audio is fed into X2I without any prior processing.
\subsection{X-to-Image} 
\begin{figure*}[htbp]
    \centering
    \includegraphics[scale=0.75]{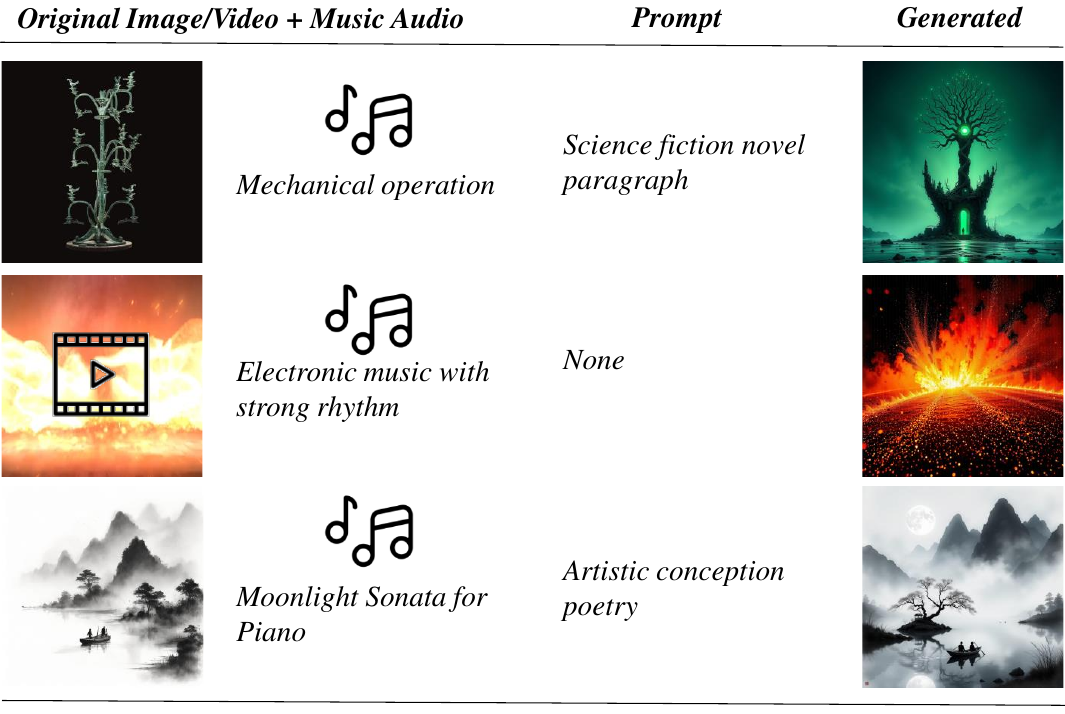}
    \caption{X-to-image generation.}
    \label{fig23}
\end{figure*}

\begin{figure*}[htbp]
    \centering
    \includegraphics[scale=1]{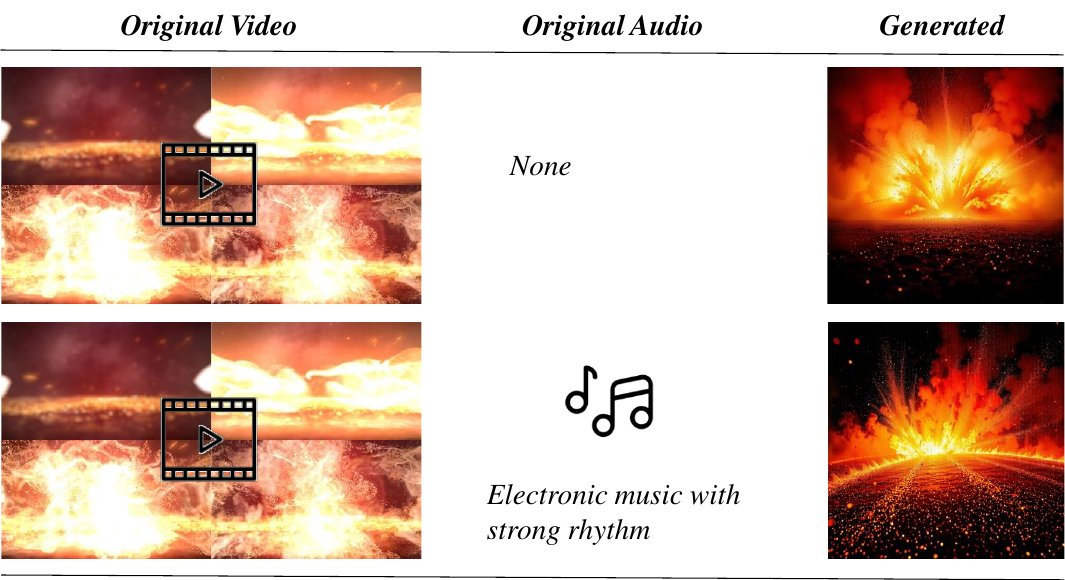}
    \caption{Images generated by combining the particle collision video with electronic music exhibit a greater sense of particle texture and oscillation compared to those created using just the video alone.}
    \label{fig24}
\end{figure*}

\begin{figure*}[htbp]
    \centering
    \includegraphics[scale=1]{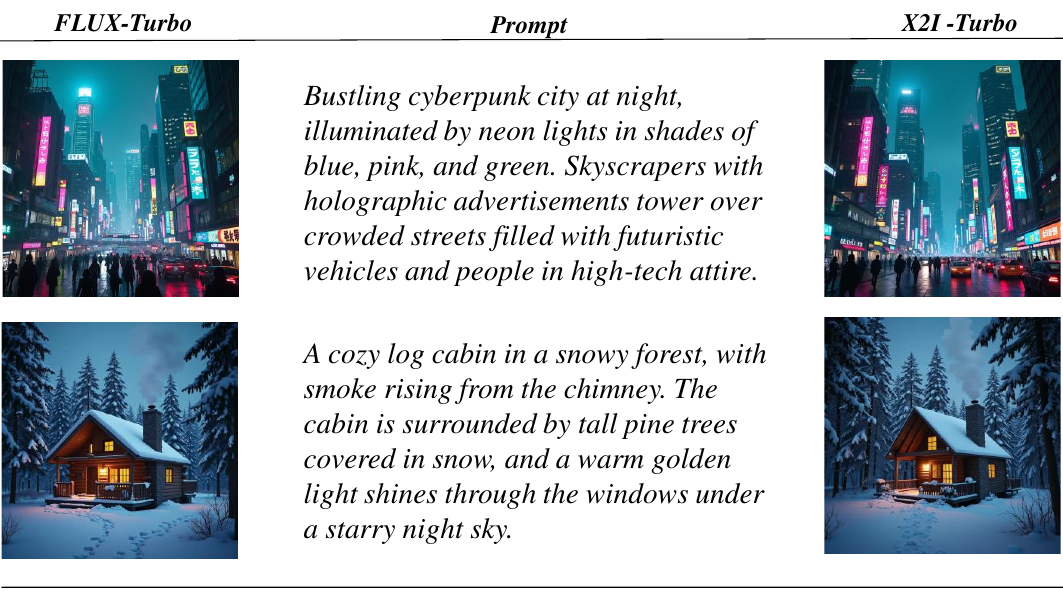}
    \caption{Comparison of the sampling acceleration model FLUX-Turbo and the X2I-Turbo obtained by replacing the text encoder of FLUX-Turbo with MLLM.}
    \label{fig25}
\end{figure*}

\begin{figure*}[htbp]
    \centering
    \includegraphics[scale=1]{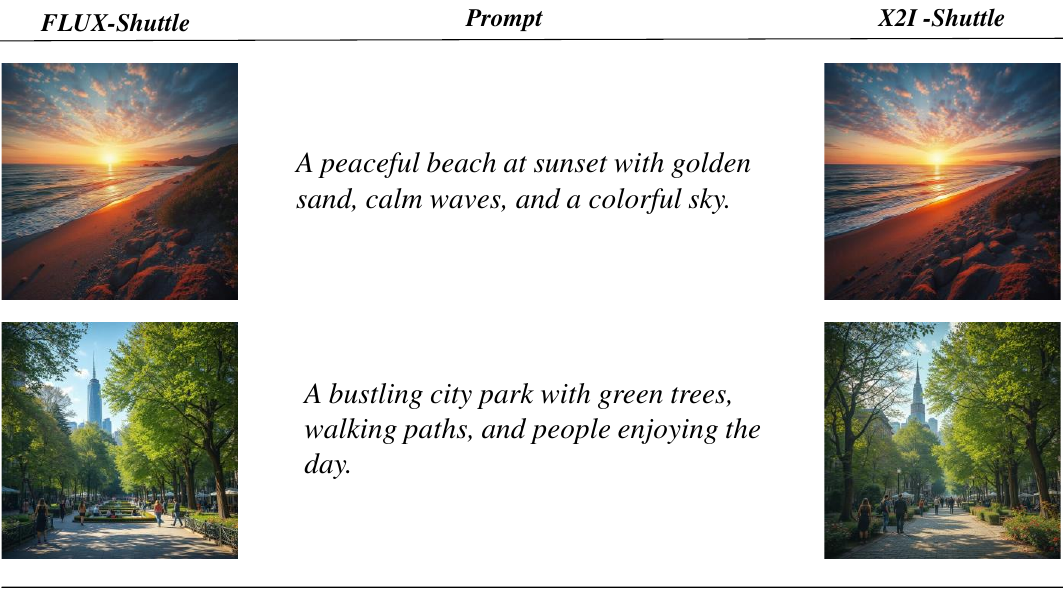}
    \caption{Comparison of the aesthetic fine-tuning model FLUX-Shuttle3.1 and the X2I-Shuttle obtained by replacing the text encoder of FLUX-Shuttle with MLLM.}
    \label{fig26}
\end{figure*}

\begin{figure*}[htbp]
    \centering
    \includegraphics[scale=1]{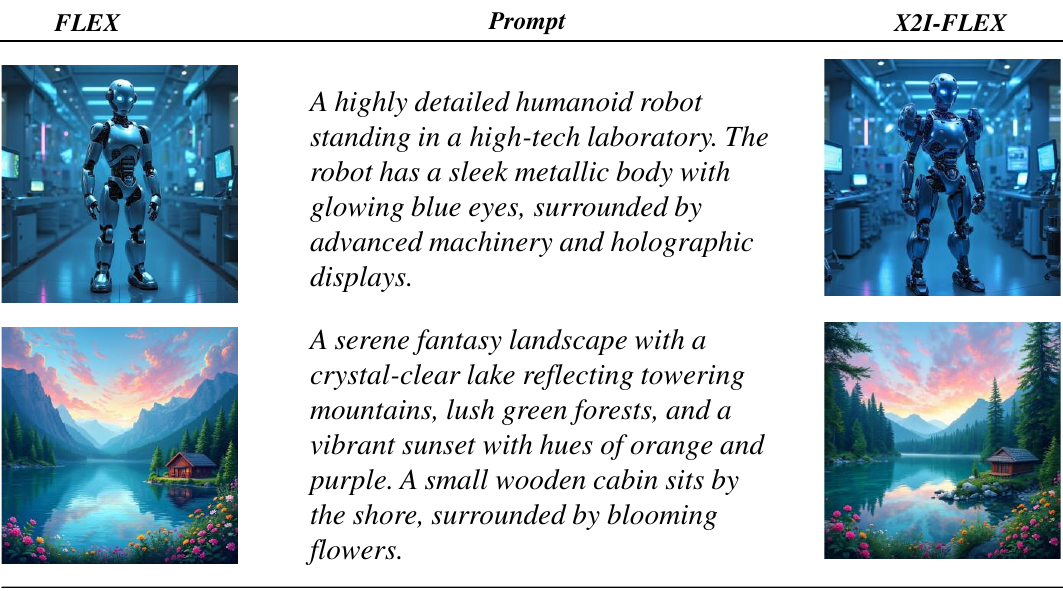}
    \caption{Comparison of the compression model FLEX and X2I-FLEX obtained by replacing the text encoder of FLEX with MLLM.}
    \label{fig27}
\end{figure*}

\begin{figure*}[htbp]
    \centering
    \includegraphics[scale=0.95]{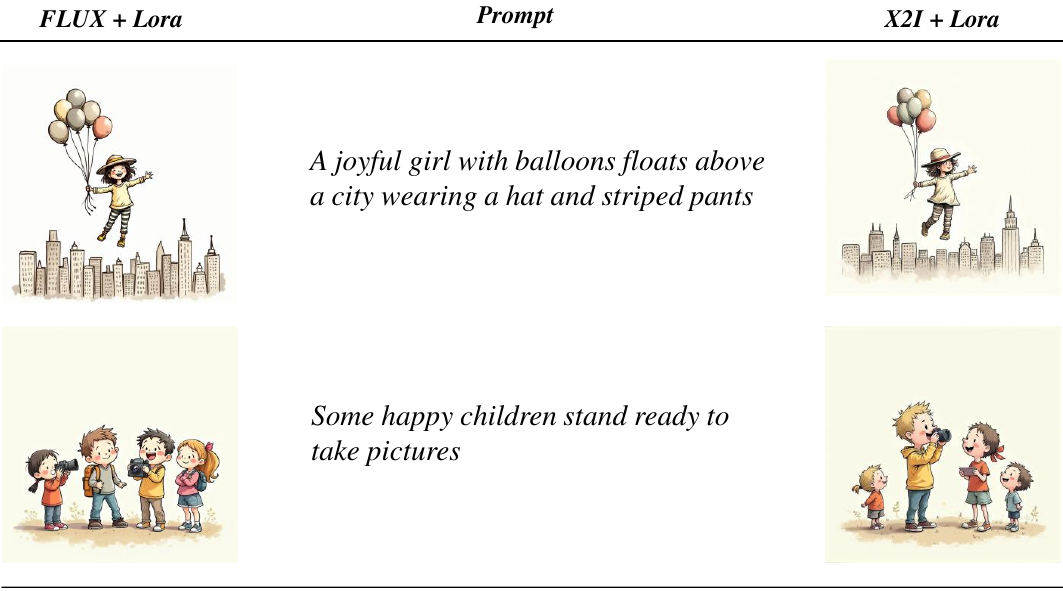}
    \caption{Comparison between FLUX.1 with children simple sketch LoRA and X2I with children simple sketch LoRA.}
    \label{fig28}
\end{figure*}

\begin{figure*}[htbp]
    \centering
    \includegraphics[scale=0.95]{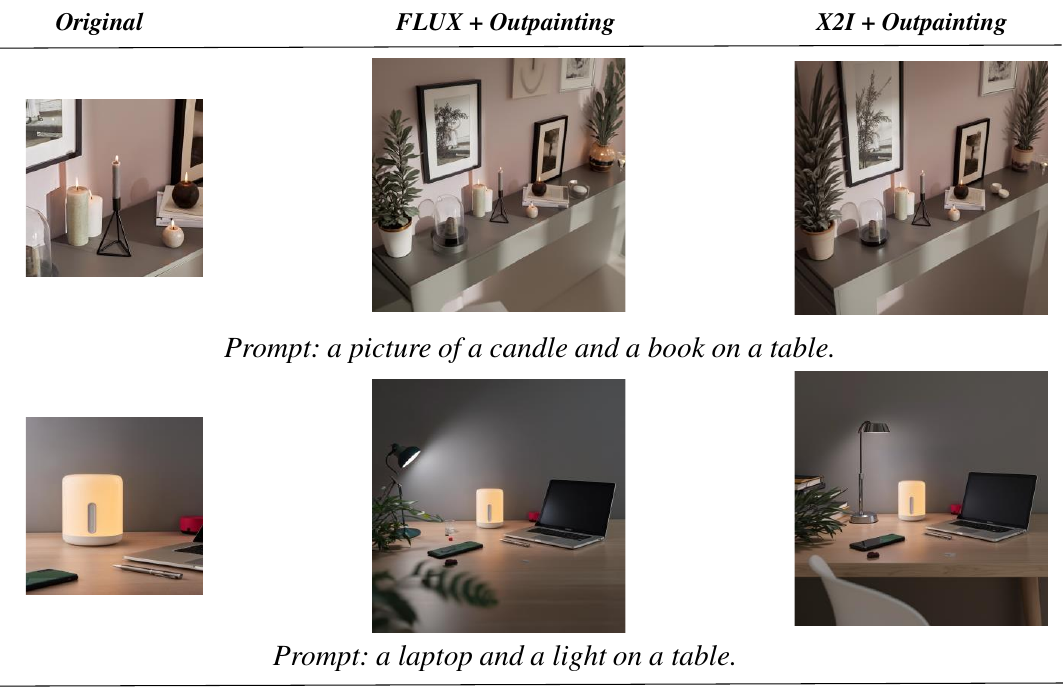}
    \caption{Comparison between FLUX.1 and X2I on the outpainting task.}
    \label{fig29}
\end{figure*}

\begin{figure*}[htbp]
    \centering
    \includegraphics[scale=1]{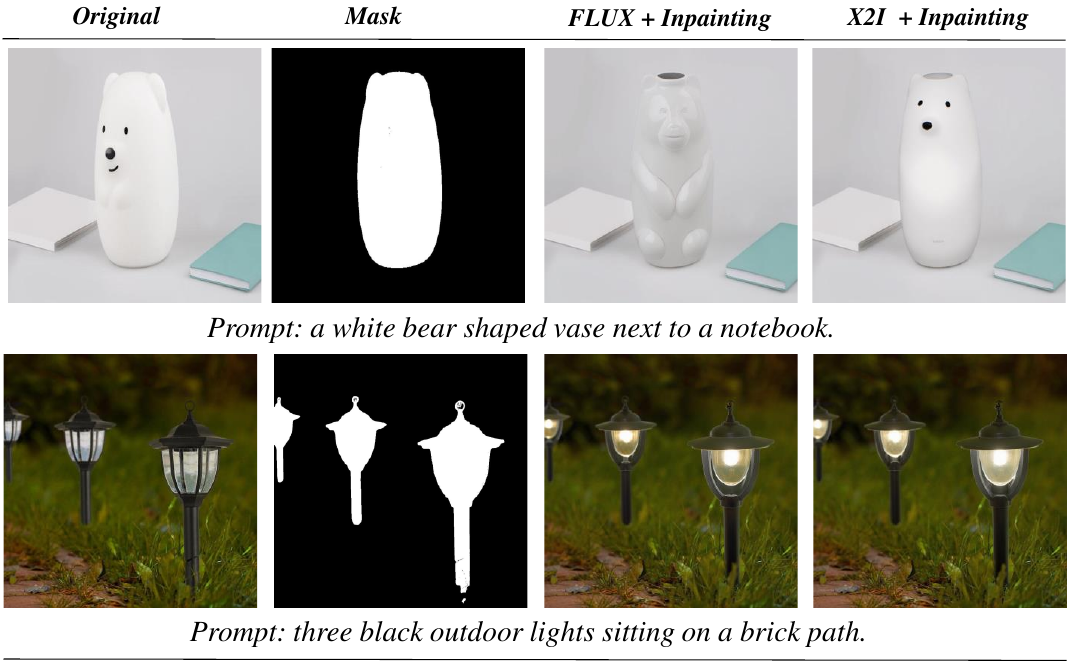}
    \caption{Comparison between FLUX.1 and X2I on the inpainting task.}
    \label{fig30}
\end{figure*}

\begin{figure*}[htbp]
    \centering
    \includegraphics[scale=1]{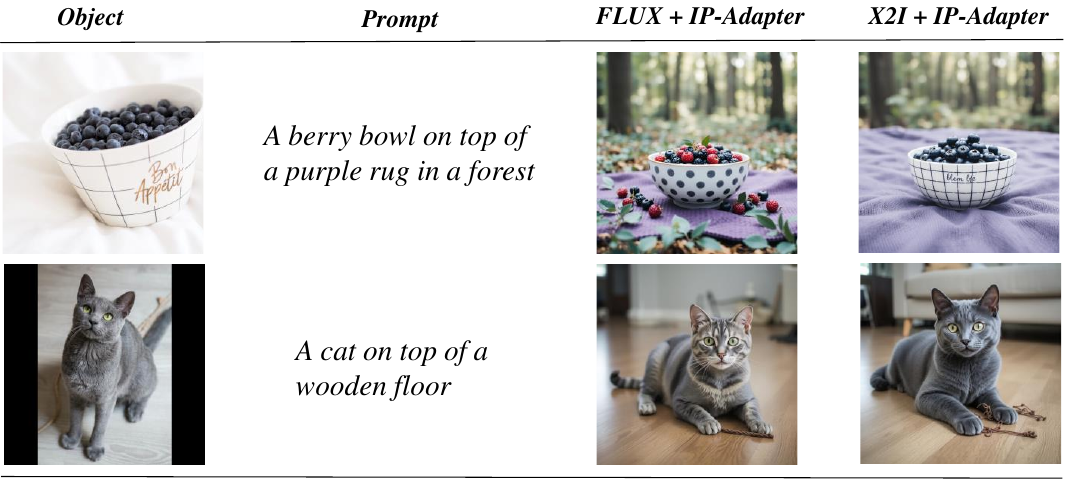}
    \caption{Comparison between Flux.1 and X2I on IP-Adapter downstream tasks. }
    \label{fig32}
\end{figure*}

\begin{figure*}[htbp]
    \centering
    \includegraphics[scale=1]{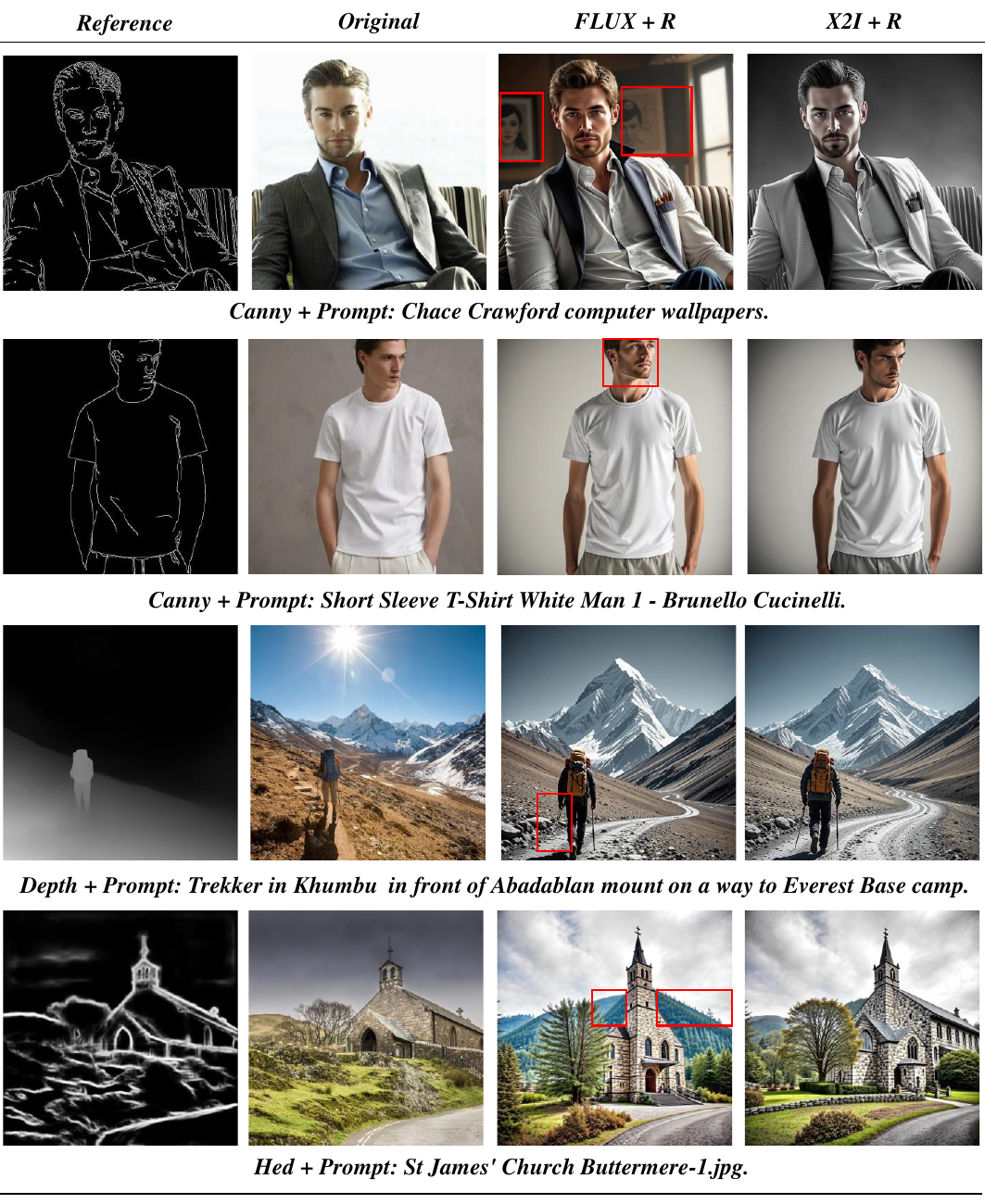}
    \caption{Comparison between Flux.1 and X2I on ControlNet tasks using canny, depth, and hed as reference maps. The annotation R indicates reference image incorporation in both FLUX.1 and X2I pipelines. Red boxes highlight X2I's superior fine-grained generation.}
    \label{fig31}
\end{figure*}

As demonstrated in \cref{fig23}, X2I can comprehend hybrid inputs combining audio, images, videos, and text prompts to generate images. The rendering of the first row has its input text description as ``In the abandoned cyberpunk ruins, a mysterious symbiotic relationship has been established between nanobots and humans. These nanobots zip back and forth through the dilapidated buildings, repairing the decaying parts of the city, while humans, by merging with the nanobots, gain extraordinary abilities. In this sci-fi world, a dazzling yet perilous new ecosystem is created, captivating people in the fantasies of the future.".
Moreover, as shown in \cref{fig24}, when the same video is used as a prompt, accompanying it with music produces distinct effects, demonstrating X2I's comprehension of multimodal prompts.

\section{Limitations}
X2I demonstrates some limitations in precise and controllable image instruction editing in a unified framework. For instance, the success rate of removing a specific element from an image is relatively low. The MLLM component of the X2I framework struggles to parse the input image into fine-grained semantic layers according to textual instructions and then output the corresponding intermediate-level features. Additionally, the LightControl have only been validated in the domain of image stylization, and further exploration is needed for more extensive image instruction editing such as addition, deletion, and modification.
Moreover, X2I have not fully exploit the multimodal understanding capabilities of MLLMs, such as summarizing long texts, logical reasoning, few-shot learning, and CoT capabilities.
Future directions should focus on more intelligent and precisely controllable visual generation. Building high-quality datasets for various image instruction editing tasks is crucial, as well as leveraging these datasets to enhance the comprehensive capabilities of X2I.

\section{Acknowledgements}
We extend our heartfelt gratitude to the dedicated researchers and scholars whose pioneering work in the fields of MLLMs and T2I generation has significantly contributed to the foundation and advancement of this study. In particular, we want to acknowledge the open-source community behind Flux.1, whose groundbreaking models have greatly facilitated our research. Your collective expertise and commitment to open science have been a constant source of inspiration and an indispensable resource for our work. Thank you for your unwavering dedication to pushing the boundaries of T2I and for fostering an environment of collaboration and innovation.






\end{document}